%% file: main.tex
\documentclass[default,iicol]{sn-jnl}

\usepackage{graphicx}%
\usepackage{tabularx}
\usepackage{multirow}%
\usepackage{amsmath,amssymb,amsfonts}%
\usepackage{amsthm}%
\usepackage{mathrsfs}%
\usepackage[title]{appendix}%
\usepackage{xcolor}%

\usepackage{textcomp}%
\usepackage{manyfoot}%
\usepackage{booktabs}%
\usepackage{algorithm}%
\usepackage{algorithmicx}%
\usepackage{algpseudocode}%
\usepackage{mdframed} %
\usepackage{listings}%
\usepackage{longtable}
\usepackage{array}
\usepackage{supertabular}
\definecolor{light-gray}{gray}{0.95}

\usepackage{subfigure}
\usepackage{array}
\newcolumntype{?}{!{\vrule width 2pt}}

\theoremstyle{thmstyleone}%
\theoremstyle{thmstyletwo}%

\theoremstyle{thmstylethree}%

\input{commands}

\begin{document}

\title[Article Title]{Semantic Anomaly Detection with Large Language Models}


\author*[1]{\fnm{Amine} \sur{Elhafsi}}\email{amine@stanford.com}

\author[1]{\fnm{Rohan} \sur{Sinha}}\email{rhnsinha@stanford.edu}

\author[1]{\fnm{Christopher} \sur{Agia}}\email{cagia@stanford.edu}

\author[1]{\fnm{Edward} \sur{Schmerling}}\email{schmrlng@stanford.edu}

\author[2]{\fnm{Issa} \sur{Nesnas}}\email{issa.a.nesnas@jpl.nasa.gov}

\author[1]{\fnm{Marco} \sur{Pavone}}\email{pavone@stanford.edu}

\affil*[1]{\orgdiv{Autonomous Systems Laboratory}, \orgname{Stanford University}, \orgaddress{\street{496 Lomita Mall}, \city{Stanford}, \postcode{94305}, \state{CA}, \country{USA}}}

\affil[2]{\orgname{Jet Propulsion Laboratory}, \orgaddress{\street{4800 Oak Grove Drive}, \city{La Cañada Flintridge}, \postcode{91109}, \state{CA}, \country{USA}}}


\input{tex/0-abstract}

\maketitle

\input{tex/1-introduction}
\input{tex/2-related-work}

\input{tex/3-methods}
\input{tex/4-experiments}
\input{tex/5-discussion}
\input{tex/6-conclusion}
\input{tex/declarations}





\begin{appendices}

\input{tex/appendices}




\end{appendices}

\bibliographystyle{unsrt}
\bibliography{references}

\end{document}

%% file: commands.tex
\usepackage{amsmath,amsfonts,bm}

\def\eqref#1{equation~\ref{#1}}

\def\1{\bm{1}}

\DeclareMathAlphabet{\mathsfit}{\encodingdefault}{\sfdefault}{m}{sl}
\SetMathAlphabet{\mathsfit}{bold}{\encodingdefault}{\sfdefault}{bx}{n}



%% file: tex/0-abstract.tex
\abstract{
As robots acquire increasingly sophisticated skills and see increasingly complex and varied environments, the threat of an edge case or anomalous failure is ever present. For example, Tesla cars have seen interesting failure modes ranging from autopilot disengagements due to inactive traffic lights carried by trucks to phantom braking caused by images of stop signs on roadside billboards. These system-level failures are not due to failures of any individual component of the autonomy stack but rather system-level deficiencies in semantic reasoning. Such edge cases, which we call \textit{semantic anomalies}, are simple for a human to disentangle yet require insightful reasoning. To this end, we study the application of large language models (LLMs), endowed with broad contextual understanding and reasoning capabilities, to recognize such edge cases and introduce a monitoring framework for semantic anomaly detection in vision-based policies. Our experiments apply this framework to a finite state machine policy for autonomous driving and a learned policy for object manipulation. These experiments demonstrate that the LLM-based monitor can effectively identify semantic anomalies in a manner that shows agreement with human reasoning. Finally, we provide an extended discussion on the strengths and weaknesses of this approach and motivate a research outlook on how we can further use foundation models for semantic anomaly detection.
}

\keywords{Semantic Reasoning, OOD Detection, Fault Monitoring}

%% file: tex/1-introduction.tex
\section{Introduction}\label{sec:intro}

Driven by advances in machine learning, robotic systems are rapidly advancing in capabilities, enabling their deployment in increasingly complex and varied scenarios. However, the infinitude of situations these robots may encounter in the real world means that we can never completely exclude the existence of rare corner cases and failure scenarios. Therefore, as well we may hope our robots generalize to novel conditions, there is an increasing need for runtime monitoring components that help issue advance warnings when a system encounters anomalies to mitigate rare failure modes. 

Since modern robots increasingly rely on learned components embedded within the autonomy stack, performance is often sensitive to so-called out-of-distribution inputs -- inputs that are dissimilar from training data \cite{KohSagawaEtAl2021, OvadiaFertigEtAl2019, GeirhosJacobsenEtAl2020, TorralbaEfros2011, SinhaSharmaEtAl2022}. Although numerous methods have been developed to detect OOD inputs \cite{SharmaAzizanEtAl2021} at the component level, such component level monitoring (e.g., detecting image classification errors) can be insufficient to prevent system-level faults. 


\begin{figure}
    \centering
    \includegraphics[width=0.47\textwidth]{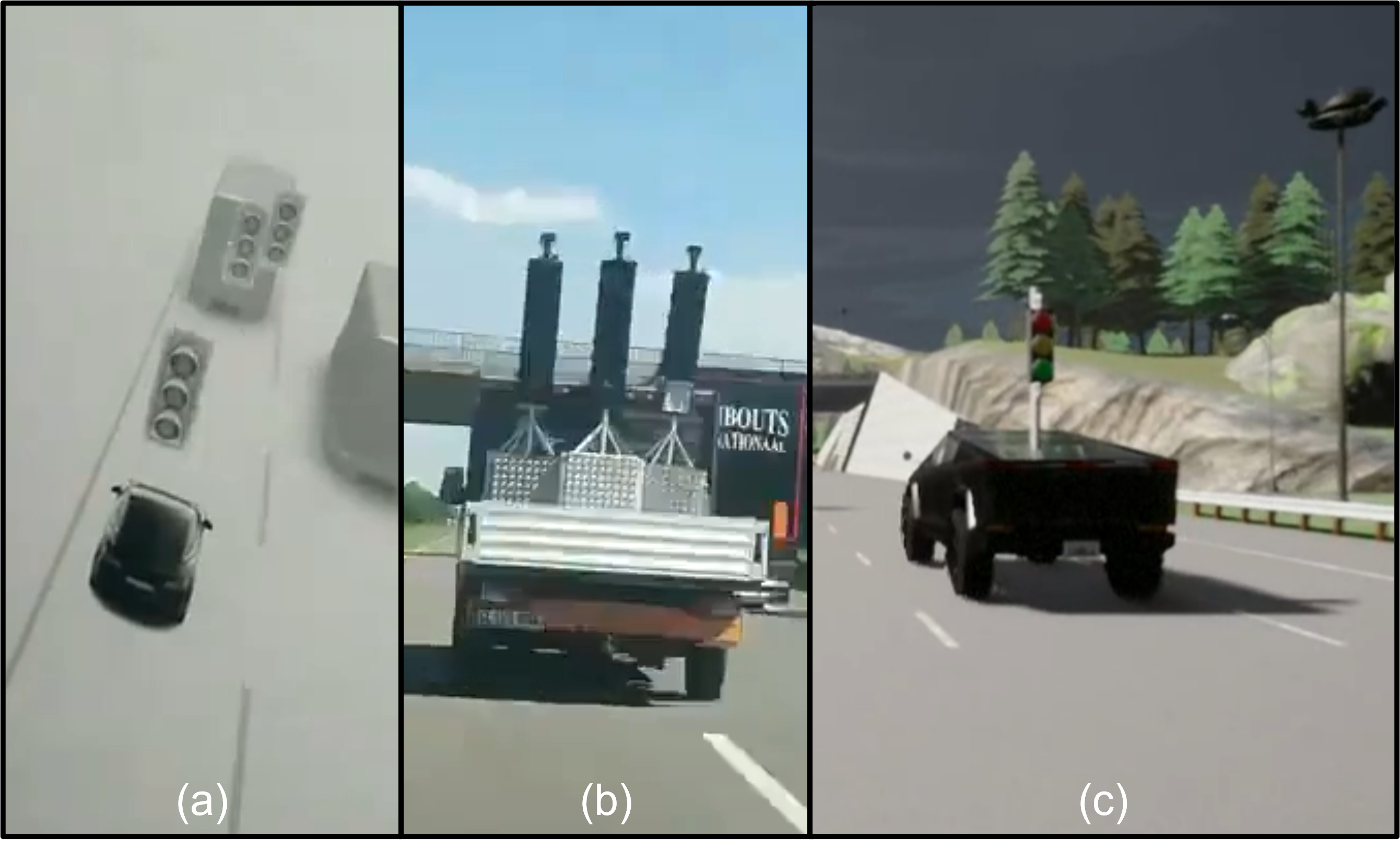}
    \caption{An example of a semantic anomaly where (a) the system visualization shows the detection of a stream of traffic lights passing through the car, (b) upon viewing the full scene it is clear that these are indeed traffic lights, just inactive in transport, and (c) a reconstruction of this scenario in the CARLA simulator which we use to demonstrate that large language models may perform the contextual reasoning necessary to detect such semantic anomalies.}
    \label{fig:tesla_carla_reconstruction}
\end{figure}

This is best exemplified by recent at-scale deployments of autonomous robotic systems, which have given rise to a steady stream of edge cases and outlier scenarios of seemingly never ending creativity. For example, among the millions of Tesla cars currently on the road, passengers have encountered disengagements due to out-of-commission traffic lights carried by a truck (see Figure~\ref{fig:tesla_carla_reconstruction})\footnote{\url{https://futurism.com/the-byte/tesla-autopilot-bamboozled-truck-traffic-lights}} and dangerous phantom braking caused by images of stop signs on roadside billboards.\footnote{\url{https://www.youtube.com/watch?v=-OdOmU58zOw}}
Similarly, recent research has shown equivalent behavior on billboards with pictures of pedestrians \cite{Gomez-DonosoEtAl2023}. These examples defy blame-assignment to a specific component: arguably, these are correct detections of traffic lights or stop signs and, at least nominally, autonomous vehicles should obey signage. Instead, these examples illustrate that it is often the context surrounding objects and their interrelations that can cause a robot to misinterpret its observations at a system-level, necessitating monitors that perform such contextual reasoning.

\begin{figure}[b!]
\begin{center}
\includegraphics[width=0.5\textwidth]{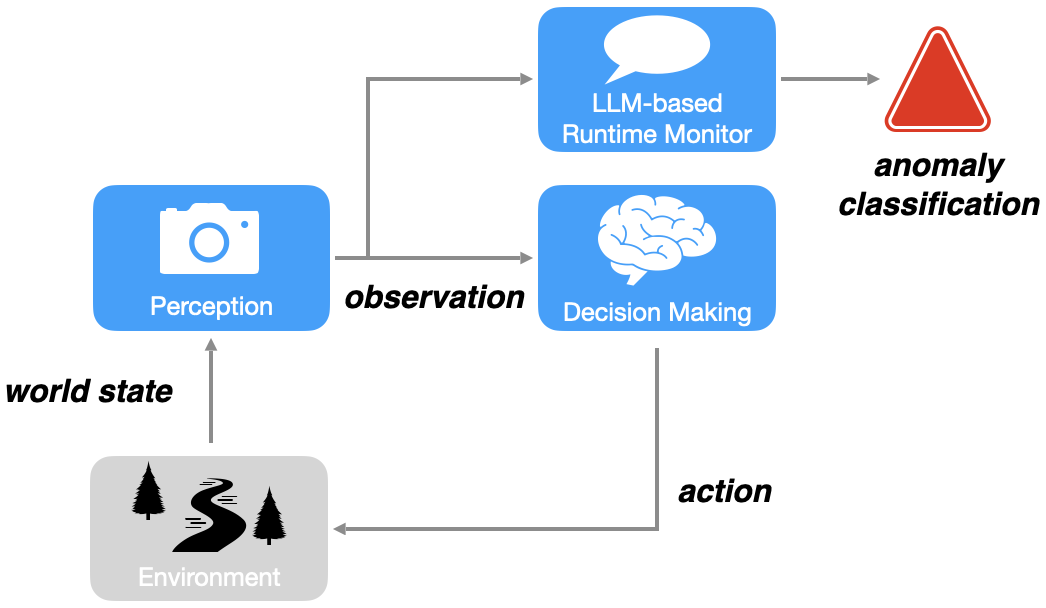}
\end{center}
\caption{Semantic anomaly detection framework. At each sampling time, the robot's observation is converted to a natural language description. The scene description is incorporated within a prompt template and passed to the LLM, which applies the necessary contextual reasoning to identify any potential semantic anomalies.}
\label{fig:overview}
\end{figure}
We refer to observations of this type as \emph{semantic anomalies}, system-level out-of-distribution (OOD) inputs that arise from an unusual or ``tricky'' combination of individually in-distribution observations. For example, they can arise because common robotics perception datasets, like BDD100K \cite{YuChen2020} for autonomous driving, do not differentiate between functioning traffic lights and those we should disregard. Moreover, even a hypothetical perception model with perfect uncertainty quantification could issue detections on images of signage on billboards without considering them OOD observations. Rather, it is their semantic irregularity---their relation to a moving truck or a roadside billboard---that causes robot failures. It is this combinatorial origin that makes these anomalies particularly difficult to guard against, for manually-programmed and learned systems alike. Conventionally, we engage a design cycle to improve the system once such a failure mode has been observed, for example by constructing targeted data augmentations \cite{Gomez-DonosoEtAl2023, VolkMullerEtAl2019}. Other approaches adapt models to changing distributions with tools from domain adaptation and continual learning \cite{WilsonCook2020, LangeAljundi2022, BanerjeeSharmaEtAl2022}. 
However, in safety-critical applications like autonomous driving, it is unacceptable to let failures occur and hope to fix them in the future. Instead, we require real time monitors that perform the necessary semantic reasoning to detect such anomalies before system failures occur. 



In this work, we study the application of foundation models trained on vast and wide-ranging data from the human experience, specifically large language models (LLMs), to provide the necessary insight to recognize semantic anomalies. We propose to do so because 1) semantic anomalies, like the traffic lights on a truck or the stop sign on a billboard, are often straightforward for a human to identify, and 2) because LLMs have demonstrated strong performance on automated reasoning tasks \cite{brown2020language, madaan2022language, weichain, kojimalarge}, so we intuitively expect the sheer scale and diversity of an LLM's training corpus to provide a broad contextual knowledge base drawn from the human experience to reason about scenarios that the robot has never experienced before.



Specifically, our contributions are three-fold. First, as illustrated in Figure \ref{fig:overview}, we introduce a monitoring framework for semantic anomaly detection for vision-based policies; this monitor applies an LLM to reason in a chain-of-thought fashion about what observed objects in a scene may result in task-relevant confusion that could cause policy errors.
Second, we instantiate and evaluate this framework in two experimental settings: 1) an autonomous driving system consisting of a perception system combined with discrete state machine logic for high-level decision making and 2) an end-to-end learned policy for object manipulation. Our experiments show that an LLM-based monitor can effectively recognize semantic anomalies, with anomaly characterizations that generally agree with human reasoning.

Finally, in light of these results, we provide an extended discussion on the success and failure modes of this approach which motivate a research outlook on how foundation models may be further adapted and utilized for the task of semantic anomaly detection.

%% file: tex/2-related-work.tex
\section{Related Work}\label{sec:relwork}

Our method leverages large language models to detect semantically anomalous scenarios that may compromise a robot's reliability. Therefore, in this section, we briefly review related work on 1) out-of-distribution and anomaly detection and 2) the use of LLMs in robotics.

\subsection{Out-of-Distribution Detection}
Commonly, deep anomaly or OOD detection methods train an auxiliary model to explicitly flag data from novel classes or data that is dissimilar from training data, aiming to detect inputs on which a DNN performs poorly \cite{SalehiMirzaeiEtAl2021, RuffKauffmanEtAl2021, YangZhouEtAl2021}. A wealth of such approaches exists. Some methods construct an OOD detection classifier on only in-distribution data \cite{hendrycksbaseline, liang2018enhancing, lin2021mood}, e.g., through so called one-class classification losses \cite{RuffVandermeulenEtAl2018}. Others directly model the training distribution \cite{ZongQiEtAl2018} and measure dissimilarity from a nominal set of values using some distance function in a latent space \cite{LeeLeeEtAl2018, denouden2018improving, MichelsAdaloglouEtAl2023}, or analyze deterioration in reconstruction errors of, e.g., autoencoders \cite{oza2019c2ae, JapkowiczMyersEtAl1995}. Alternative to direct OOD classification, other methods seek to improve the often poorly calibrated uncertainty scores of neural networks on OOD data \cite{AbdarPourpanahEtAl2021}. Many are based on a Bayesian viewpoint and emphasize tractable approximations to the Bayesian posterior via,  e.g., deep ensembles \cite{LakshminarayananPritzelEtAll2017}, Monte-Carlo dropout \cite{GalZoubin2016}, or sensitivity-based Laplace approximations \cite{SharmaAzizanEtAl2021, ritter2018scalable}.
Alternatively, several approaches construct architectures, loss functions, or training procedures to design networks that output high predictive uncertainty outside the training data \cite{AminiSchwartingEtAl2020, OsbandZhenEtAl2023, LiuLinEtAl2020, AbdarPourpanahEtAl2021}. 

Beyond OOD detection for individual models, anomaly detection methods have been developed that make comparisons between multiple related models, leveraging domain knowledge on expected relationships to flag potential faults.
For example, \cite{AntonanteSpivakEtAl2021, LeeTanEtAl2021} demonstrate that we can detect perception errors by checking consistency across sensing modalities (e.g., camera, LiDAR) and over time. In addition, other methods apply supervised learning on examples of successes and failures based on ground truth information or domain specific supervisory signals \cite{DaftryZengEtAl2016, RabieeBiswas2019}. 

While these existing methods for OOD/anomaly detection have proven useful in increasing the reliability of downstream decision-making \cite{RichterRoy2017, McAllisterKahn2019}, their application comes with two limitations. First, these existing approaches cannot reason about robot failure modes that arise from semantically anomalous scenarios as defined in the preceding section. Instead, they only aim to detect anomalies correlated with inference errors or quantify uncertainty in the perception system and its predictions. That is, they monitor the correctness of individual components (even in the aforementioned case of multimodal perception), rather than the aggregate reasoning of an autonomy stack. Second, a limitation of existing OOD/anomaly detection methods is that we cannot immediately apply them as general purpose diagnostic tools: they require access to the model's training data, data of failures, or domain-specific knowledge. Instead, we propose the use of LLMs to perform zero-shot reasoning over possible reasoning errors that may be made about an observed scene with respect to a task description. As a result, our method is a general-purpose diagnostic tool that we can directly apply to many tasks without requiring access to training data or task-specific component redesign. 


It is commonly understood that unexpected inference errors often occur at test time because models overfit to spurious correlations that may hold on a particular dataset, but not in general \cite{GeirhosJacobsenEtAl2020, TorralbaEfros2011}. To overcome spurious correlations, recent work in domain generalization (that is, increasing robustness to unknown distributional shifts as defined in \cite{GulrajaniLopes2021}) has sought to disentangle the causal relations between inputs and outputs \cite{ArjovskyBottouEtAl2020}. Furthermore \cite{HaanJayaraman2019} apply similar principles of causal inference to learn to avoid semantically motivated reasoning errors of end-to-end learned policies. However, doing so requires sufficient natural interventions, and thus observations of semantic anomalies, to prune spurious correlations. By definition, such anomalies are extremely rare, so zero-shot approaches to reason about semantic anomalies are necessary to avoid failures. To more structurally enable semantic reasoning over a scene, some recent work has developed methods for perception that construct scene representations that capture the relations between objects and the environment as 3D Scene Graphs \cite{RosinolVioletteEtAl2021} and \cite{ChenHuEtAl2022} has applied LLMs to classify room type from objects contained in a 3D Scene Graph. We further investigate the ability of LLMs to flexibly interpret relational scene descriptions. A key difference is that we investigate reasoning from the semantic context of a downstream decision making task rather than considering a classification task in isolation. 

\subsection{Large Language Models in Robotics}

Thanks to their broad and general purpose knowledge, LLMs are becoming a ubiquitous part of robotic systems that can be adapted to a range of downstream applications. LLMs have been used for various purposes such as autonomous navigation \cite{shah2023lm}, long-horizon planning \cite{huanginner, lin2023text2motion}, \cite{brohan2023can} and policy synthesis \cite{liang2022code}. LLM embeddings have also been used to achieve language-conditioned control as in \cite{cui2022can}. The common approach among these works is the integration of LLMs within a broader robotics subsystem (e.g., perception, planning, etc.) to improve functionality or make the subsystem amenable to tasks that require linguistic reasoning. In contrast, we seek to leverage LLMs as a ``semantic reasoning" module that monitors the observations that an independent robot (or subsystem thereof) encounters. 

A line of works demonstrate that LLMs are avid reasoners in few-shot settings \cite{brown2020language}, \cite{madaan2022language}, particularly when LLMs are prompted to explain the logical steps underpinning their response \cite{weichain} Similar prompting strategies have been developed to elicit zero-shot reasoning from LLMs \cite{kojimalarge}. In many real-world cases, robot failures may be predictable from semantically anomalous conditions; conditions that can be identified with such LLM prompting strategies before the failure occurs.

Large pretrained models have also been used at the intersection of computer vision and language, as in Socratic Models \cite{zeng2022socratic} wherein large pretrained models are composed in a zero-shot fashion to perform tasks such as video understanding and image captioning. Multimodal models such as Flamingo \cite{alayracflamingo}, BLIP \cite{li2022blip} and PaLM-E \cite{driess2023palme} have also demonstrated impressive performance on image and video understanding tasks. However, characterizing the suitability of a scene for a particular downstream task is an unsolved problem.

%% file: tex/3-methods.tex
\section{LLM-Based Semantic Anomaly Detection}\label{sec:approach}

In this study, we propose a monitoring framework that leverages an LLM-based module to supervise a robot's perception stream while it operates, identifying any semantic anomalies that it may encounter. At some sampling frequency, the monitor examines the robot's visual observations and transforms them into textual descriptions. These scene descriptions are then integrated into an LLM prompt that aims to identify any factors that could give rise to a policy error, unsafe behavior, or task confusion. We emphasize that the level of task and system specificity of the monitor is determined by the context provided in the prompt. In particular, the role of the LLM-based monitor is to indicate potential factors in the visual observations that could contribute to confusion, but it is neither explicitly grounded in the system's capabilities or deficiencies nor does it access the system's internal processes. Figure \ref{fig:overview} provides an illustration of this framework.

An essential component of this framework is the conversion of visual information into natural language descriptions. Our method is agnostic to the approach used for visual-to-text conversion, and various techniques can be utilized ranging from classical image processing to vision-language models (VLMs) depending on the task requirements. In this work, we employ an open vocabulary object detector \cite{minderer2022simple}, which describes the objects present in a scene along with associated context.

To enhance the LLM's ability to reason over the scene descriptions, we employ prompting strategies that leverage current best practices in prompt engineering. Specifically, we employ few-shot prompting \cite{brown2020language, madaan2022language} and chain-of-thought reasoning \cite{kojimalarge} in our instantions of this framework. Few-shot prompting provides examples to prime the LLM's reasoning. These examples highlight common pitfalls and reasoning errors that the robot may encounter during operation, and serve to guide the LLM towards more robust decision-making. Chain-of-thought reasoning adds specific task- and system-specific structure to the reasoning.

We describe and evaluate two instantiations of this framework in the subsequent section.






%% file: tex/4-experiments.tex
\section{Experiments}\label{sec:expts}

In this section, we assess an LLM's ability to identify situations that could potentially result in semantic failures through several experiments, and provide insights into the strengths and limitations of the approach. We perform experiments in an autonomous driving setting with a reasoning-based policy and in a manipulation setting with an end-to-end learned visuomotor policy.

We first present the autonomous driving experiments, which are primarily motivated by the aforementioned real-world Tesla failure cases induced by semantic anomalies. These experiments instantiate a complete implementation of the proposed anomaly detection framework and demonstrate its effectiveness in recognizing semantic anomalies. To confirm the novel detection capabilities, we also compare against two existing OOD baselines founded on different detection principles, and indeed demonstrate that these methods complement each other in discerning different types of anomalies. 

Second, we present the manipulation experiments to evaluate the utility of an LLM-based monitor when the system of interest is controlled by a purely learned (``black box'') policy. These experiments demonstrate that, although the LLM-based monitor still tends to produce anomaly characterizations that align with human expectations, learned policies can often fail in erratic or inexplicable ways, which limits the versatility of our approach.


\subsection{Reasoning-Based Policy}

\subsubsection{Experimental Setup}
We conducted autonomous driving experiments in the CARLA simulator \cite{dosovitskiy2017carla}, a platform which provides realistic 3D environments for testing and validating autonomous driving systems. The goal of these experiments is to demonstrate that the LLM-based monitor can effectively identify autonomous driving semantic edge cases in an end-to-end manner, without requiring additional training or fine-tuning. We designed five classes of scenarios for our experiments:

\begin{enumerate}
    \item Nominal Stop Sign Interaction: The autonomous vehicle encounters stop signs at intersections where it is required to make a complete stop before proceeding.
    \item Nominal Traffic Light Interaction: The autonomous vehicle encounters traffic lights and must obey their signals.
    \item Anomalous Stop Sign Interaction: In these scenarios, we introduce a billboard or poster with a stop sign as part of the graphic along the vehicle's route, in order to simulate situations where the vehicle's perception system might mistake a non-functional stop sign for a real one and cause the vehicle to stop unnecessarily.
    \item Anomalous Traffic Light Interaction: In these scenarios, the autonomous vehicle encounters a truck transporting a traffic light along its route, to simulate situations where the autonomous vehicle may be influenced by a non-functional traffic light.
    \item Strange Objects: In these scenarios, the autonomous vehicle encounters an entity unexpected to be seen on the road in an adjacent lane. We introduce one of five objects for each episode chosen from an airplane, a boat, an elephant, a robot, and a train.
\end{enumerate}

We handcrafted 10, 10, 16, 18, and 15 cases for the respective scenario classes which were evenly distributed over all public CARLA maps to encompass diverse driving environments representing suburban, urban, rural and highway settings.  

To simulate the autonomous vehicle's driving behavior, we developed a finite-state-machine-based planner designed to follow a route while obeying stop sign and traffic light signals. Perception is achieved through a single forward-facing RGB camera fixed on the vehicle. Stop signs and traffic lights native to the CARLA maps are detected by querying privileged simulator information once they enter the vehicle's field of view. The corresponding stopping regions are also retrieved with the detections. To determine the active traffic light color, we developed a simple approach that involved masking green, yellow, and red pixels on the traffic light patch and identifying the most prominent color.

To test the vehicle's performance in the presence of anomalies, we employed YOLOv8 \cite{Jocher_YOLO_by_Ultralytics_2023} to detect any stop signs or traffic lights that may exist in addition to those present in the environment by default. Detected stop signs cause the vehicle to come to a temporary stop at the point on its route nearest the detection. Similarly, traffic light detections can influence the vehicle's behavior based on the classified active color.

In our simulations, all edge cases caused the vehicle to stop in unsafe locations, such as in the middle of intersections or on the freeway, demonstrating that the detection of such semantic edge cases meaningfully improves autonomous driving safety.

\subsubsection{Anomaly Detection Methodology}

\begin{figure*}[h!]
\begin{center}
\includegraphics[width=0.82\textwidth]{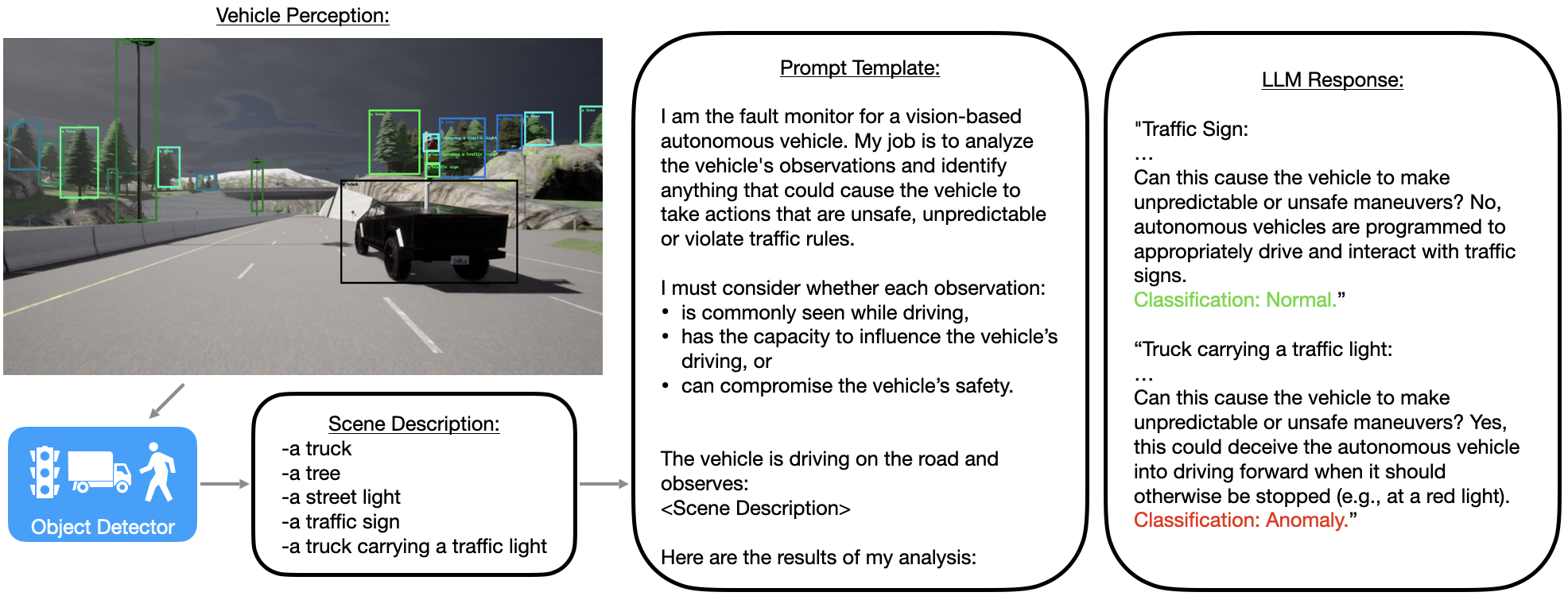}
\end{center}
\caption{An overview of the autonomous driving semantic anomaly detection methodology. Vehicle perception is sampled at a frequency of 2 Hz and passed through an open vocabulary object detector to generate a scene description. The scene description is incorporated within a prompt template, which is used to query the LLM. Note that the prompt template and response depicted are respectively paraphrased from the full prompt (Appendix \ref{sec:carla_prompt}) and excerpted from the full output.}
\label{fig:carla_exp_methodology}
\end{figure*}

To detect anomalies, we generated prompts from the autonomous vehicle's observations and fed them to an LLM. Observations were evaluated at a frequency of 2 Hz. In tandem, we also evaluate two baseline OOD detection methods.

\textbf{LLM Anomaly Detection (Ours):} First, the observed images were parsed by the OWL-ViT open vocabulary object detector to yield a scene description, consisting of the detected objects along with relevant context. To generate descriptive labels, we constructed a vocabulary by enumerating a diverse set of objects ranging from vehicle types to scenery and pairing them with descriptive predicates (e.g., ``near the road," ``on the bridge," etc.). The scene description was then integrated into a prompt template designed to elicit a chain-of-thought reasoning style to identify whether any element of the scene might cause erroneous, unsafe, or unexpected behavior. Finally, the \texttt{text-davinci-003} LLM analyzed the prompt to identify any potential semantic anomalies present in the observation. An illustration of this process is depicted in Figure \ref{fig:carla_exp_methodology}.

\textbf{SCOD \cite{sharma2021sketching} (Baseline):} As a baseline, we apply the SCOD OOD detector to the object detection model used in the autonomy stack. SCOD wraps an existing model to produce a scalar anomaly signal that reflects a model's epistemic uncertainty in a task-relevant manner. Specifically, since SCOD is based on the Laplace approximation of the Bayesian posterior over models given their training data, the SCOD metric essentially functions as a measure of confidence in a model's predictions on unseen and possibly OOD inputs. To construct the SCOD metric in an efficient manner, we wrap the detection head (which outputs a categorical distribution over classes in a bounding box) of the object detector, similar to the practice of wrapping the final layers of a model advocated in \cite{sharma2021sketching}. SCOD requires a-priori processing of local curvature information of predictions on the training data, so we instantiate SCOD with the MS-COCO dataset used to train the object detector. 

\textbf{Mahalanobis Distance (Baseline):} As a second baseline, we compare against a Mahalanobis distance metric, a simple and typical anomaly detection method that fits a Gaussian mixture model (GMM) to the feature embeddings of nominal, in-distribution inputs and evaluates the likelihood of test inputs under the nominal distribution. Specifically, we perform a dimensionality reduction on the feature embeddings of the object detector's ResNet backbone and fit a GMM in reduced coordinates using the images from the nominal driving scenarios. Since this approach explicitly models the nominal input distribution, we can expect it to issue warnings on inputs that look visually distinctive from nominal observations.    

\subsubsection{Results}

\begin{table*}[]
    \caption{Results of our LLM-based monitor on finite state machine-based autonomous vehicle stack in the CARLA simulator. The first two rows report the total number of semantic anomalies and the total number of nominal observations in the dataset. We distinguish between baseline episodes without any anomalies (nominal episodes) and the test cases where a semantic anomaly is visible for a short duration (anomalous episodes), each of which is further split into episodes where the AV encounters a particular type of interaction. A true positive (TP) occurs when a semantic anomaly is in view and the monitor issues an alert, a false negative (FN) occurs when the anomaly is in view but the monitor does not detect it. A true negative (TN) occurs when no anomalies are in view and the LLM does not detect anomalies, a false negative (TN) occurs when an anomaly is in view and the monitor does not detect so. We report the rate for each of these cases respectively.}
    \centering
    \begin{tabularx}{\textwidth}{c|X|X?X?X|X|X?X?X?}
    \hline
         & \multicolumn{3}{c}{\textbf{Nominal Episodes}} & \multicolumn{4}{|c|}{\textbf{Anomalous Episodes}} & \multirow{2}{4em}{\textbf{Total}} \\
         & Stop Signs & Traffic Lights & \textbf{Total} & Stop Signs & Traffic Lights & Strange Objects &\textbf{Total} &\\
         \hline
         Semantic Anomalies & 0 & 0 & 0 & 16 & 19 & 15 & 50 & 50 \\
         Nominal Observations & 309 & 494 & 803 & 248 & 197 & 337 & 782 & 1585 \\ \hline 
         TPR & N/A & N/A & N/A & 0.94 & 0.84 & 1.0 & 0.92 & 0.92 \\
         FNR & N/A & N/A & N/A & 0.06 & 0.16 & 0.0 & 0.08 & 0.08 \\
         TNR & 0.91 & 0.54 & 0.68 & 0.86 & 0.80 & 0.93 & 0.88 & 0.78  \\
         FPR & 0.09 & 0.46 & 0.37 & 0.14 & 0.20 & 0.07 & 0.12 & 0.22\\
    \hline
    \end{tabularx}
    
    \label{tab:carla-results}
\end{table*}

\begin{table*}[]
    \caption{Results of our LLM-based monitor on finite state machine-based autonomous vehicle stack in the CARLA simulator \textbf{using ground truth scene descriptions}.}
    \centering

    \begin{tabularx}{\textwidth}{c|X|X?X?X|X|X?X?X?}
    \hline
         & \multicolumn{3}{c}{\textbf{Nominal Episodes}} & \multicolumn{4}{|c|}{\textbf{Anomalous Episodes}} & \multirow{2}{4em}{\textbf{Total}} \\
         & Stop Signs & Traffic Lights & \textbf{Total} & Stop Signs & Traffic Lights & Strange Objects &\textbf{Total} &\\
         \hline
         Semantic Anomalies & 0 & 0 & 0 & 16 & 19 & 15 & 50 & 50 \\
         Nominal Observations & 309 & 494 & 803 & 248 & 197 & 337 & 782 & 1585 \\
    \hline 
     TPR & N/A & N/A & N/A & 1.0 & 1.0 & 1.0 & 1.0 & 1.0 \\
     FNR & N/A & N/A & N/A & 0.00 & 0.00 & 0.00 & 0.00 & 0.0 \\
     TNR & 0.97 & 0.85 & 0.90 & 0.96 & 0.94 & 1.00 & 0.95 & 0.92  \\
     FPR & 0.03 & 0.15 & 0.10 & 0.04 & 0.06 & 0.00 & 0.05 & 0.08\\
    \hline
    \end{tabularx}
    
    \label{tab:carla-results-groundtruth}
\end{table*}

\begin{table*}[]
    \caption{Baseline comparison results for semantic anomaly detection using the LLM monitor, SCOD, and the Mahalanobis distance on the FSM-based AV stack in the CARLA simulator. As in Table \ref{tab:carla-results}, the TPR and TNR indicate the true positive and negative rates of an OOD detectors' ability to detect semantic anomalies.}
    \centering
    \begin{tabularx}{\textwidth}{cX|X?X?X?X|X|X?X?X?}
    \hline
         & & \multicolumn{3}{c}{\textbf{Nominal Episodes}} & \multicolumn{4}{|c|}{\textbf{Anomalous Episodes}} & \multirow{2}{4em}{\textbf{Total}} \\
         & & Stop Signs & Traffic Lights & \textbf{Total} & Stop Signs & Traffic Lights & Strange Objects &\textbf{Total} &\\
         \hline
         \multirow{2}{4em}{\textbf{LLM}} & TPR & N/A & N/A & N/A & 0.94 & 0.84 & 1.0 & 0.92 & 0.92 \\
         & TNR & 0.91 & 0.54 & 0.68 & 0.86 & 0.80 & 0.93 & 0.88 & 0.78  \\
         \hline
         
         \multirow{2}{4em}{\textbf{SCOD}} & TPR & N/A & N/A & N/A & 0.13 & 0.63 & 0.07 & 0.30 & 0.30 \\
         & TNR & 0.94 & 0.96 & 0.95 & 0.89 & 0.97 & 0.89 & 0.91 & 0.93  \\
         \hline
         
         \multirow{2}{4em}{\textbf{Mahal.}} & TPR & N/A & N/A & N/A & 0.31 & 0.47 & 0.67 & 0.48 & 0.48 \\
         & TNR & 0.93 & 0.96 & 0.95 & 0.70 & 0.77 & 0.50 & 0.63 & 0.79  \\
         \hline
    \end{tabularx}
    
    \label{tab:carla-results-baselines}
\end{table*}

\begin{table}
    \caption{Results on detection of errors in the outputs of the object detector used in the FSM-based AV stack: A true positive indicates that a warning was issued when the detector produces an erroneous bounding-box or misclassifies an object. A false positive indicates that a warning was issued when the object detector correctly identified all objects relevant to the driving task. We only consider the nominal episodes here to avoid any ambiguity on whether e.g., a detection of an image of a stop sign on a billboard as a stop sign qualifies as a perception error.}
    \label{tab:inference}
    \centering
    \begin{tabularx}{\linewidth}{c|X|X|X|}
    \hline
     & \multicolumn{2}{c|}{\textbf{Nominal Episodes}} \\
    Method & TPR & FPR\\
    \hline
    LLM & 0.09 & 0.42  \\
    SCOD & 0.80 & 0.46  \\
    Mahalanobis & 0.39 & 0.55  \\
    \hline
    \end{tabularx}
\end{table}

We assess the performance of our semantic anomaly detection methodology by comparing the LLM-based monitor's anomaly detections with ground truth labels determined based on the visibility of the anomaly. To compute these values, we first split each episode into distinct time intervals corresponding to periods where the anomaly either is or is not in view. If the LLM yields an anomaly classification for any timestep during which the anomaly is in view, we treat the entire interval as a single true positive (TP). Conversely, if the anomaly is not detected at any point in the interval, we treat the interval as a false negative (FN). We count true negatives (TNs) and false positives (FPs) on a per timestep basis when an anomaly is not in the field of view. We present the results for the four different scenario types in Table \ref{tab:carla-results}.

We note that the LLM demonstrates strong performance across the anomalous stop sign and traffic light, and nominal stop sign scenarios. The results indicate a high true positive rate (TPR) and low false negative rate (FNR) in the anomalous scenarios with relatively lower rates of misdetections. In the nominal stop sign episodes, the LLM demonstrates a low false positive rate (FPR). However, we interestingly see an elevated FPR in the nominal traffic light episodes. However, taken altogether, the detections indicate strong reason capabilities in realistic nominal and anomalous scenarios.

\begin{figure}[]
\begin{center}
\includegraphics[width=0.5\textwidth]{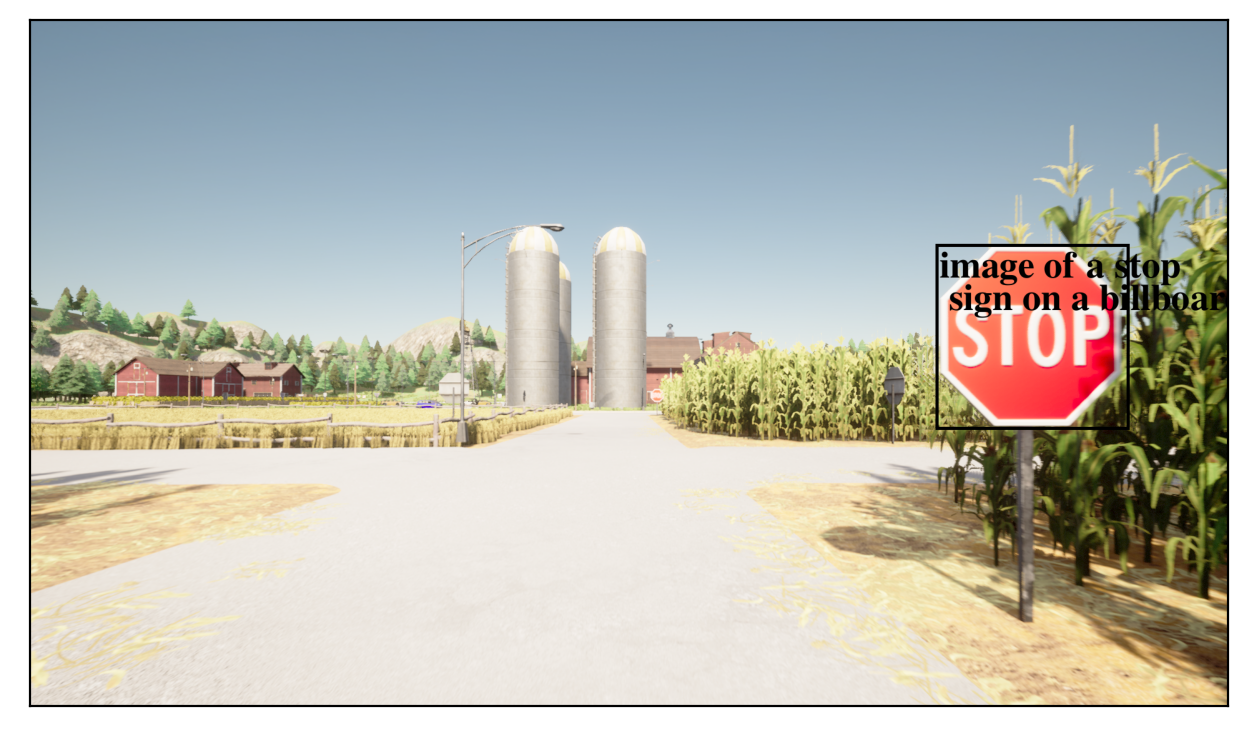}
\end{center}
\caption{Example of an object detection error. The stop sign in the scene is mistaken for an ``image of a stop sign on a billboard."}
\label{fig:carla_object_misdetection}
\end{figure}

We noted that CARLA's visual fidelity, though of high quality, differs from that of the real-world images on which OWL-ViT was trained. As a result, we found that the object detector would occasionally misclassify objects or hallucinate unseen objects altogether, possibly due to lighting conditions or artifacts in the simulated visuals. One common error that we observed was the detection of objects with the predicate ``on a billboard," despite the fact that no billboards were in the scene, an example of which is illustrated in Figure \ref{fig:carla_object_misdetection}. We hypothesize that this issue is a result of background lighting effects creating the illusion of a flat background. To demonstrate that the object detector's errors were the primary source of anomaly misclassifications we also present anomaly detection performance with ground truth scene labels in Table \ref{tab:carla-results-groundtruth}. 

We emphasize the LLM-based monitor's unique ability to detect semantic anomalies by comparing its performance to two OOD detection methods: SCOD, the method proposed by \cite{sharma2021sketching}, and a Mahalanobis distance-based approach. We apply these OOD detectors to the DETR object detection model \cite{carion2020end}\footnote{Although we use YOLOv8 \cite{Jocher_YOLO_by_Ultralytics_2023} in our vehicle planner, we find that DETR yields similar performance. We apply the baselines to DETR as it is trained on the same dataset as YOLOv8, though its architecture is more amenable to applying traditional OOD detectors.}. Since these baselines produce real-valued OOD metrics, we tune the threshold at which we signal an anomaly as the 95\% quantile of the OOD signal on observations from the nominal driving scenarios. This ensures a false positive rate of 5\%. As shown in Table \ref{tab:carla-results-baselines}, we observe that the LLM-based monitor significantly outperforms both SCOD and the Mahalanobis distance at the CARLA semantic anomaly detection task. 

We can explain the poor semantic anomaly detection performance of our baselines by examining the differences in their underlying principles: the SCOD metric aims to detect when the perception system produces erroneous predictions and the Mahalanobis distance functions as a measure of distinctiveness in an observation's visual appearance with respect to nominal data. Since semantic anomalies typically consist of unusual combinations of commonly seen, in-distribution elements (e.g., stop sign imagery or traffic lights on trucks), the resulting failure modes can occur even when the AV's perception model confidently produces correct predictions or when observations appear ``normal'' at a pixel-level. In contrast, the LLM excels at identifying these tricky scenarios. 

Nonetheless, we point out that semantic anomaly detection and traditional OOD detectors are not mutually exclusive tools. In fact, we highlight the complementarity of our LLM-based monitor and the baseline OOD detectors with an additional set of results. In Table \ref{tab:inference}, we contrast the ability of our method and the baselines to detect when the AV's perception system makes erroneous object detections, a meaningful task for a component-level runtime monitor. Here, we tuned the baselines' detection thresholds so that the false positive rate was roughly 50\% to show their predictive ability of detecting perception inference errors. We see that the LLM is completely unable to detect perception errors, whereas SCOD is able to do so. The Mahalanobis baseline also demonstrates some capacity to detect perception errors, although here it does not perform as well as SCOD. 


Finally, it is important to emphasize that while these principles are complementary, they are not disjoint: Table \ref{tab:carla-results-baselines} shows that SCOD has some capacity to the detect the traffic light on a truck, and the Mahalanobis distance has some capacity to detect strange objects. Indeed, at a qualitative level, we observed that the Tesla Cybertruck asset used in the traffic light anomalies was often not detected by the AVs perception system, hence this semantic anomaly often also corresponds with perception errors. In addition, the assets used for the strange objects appear quite cartoonesque, so they may appear quite visually distinct from the real photos of elephants, boats, and airplanes in the MS-COCO dataset or the realistic objects in the nominal CARLA scenarios.




\subsection{Learned Policy}
\subsubsection{Experimental Setup}

We applied our LLM-based runtime monitor to a manipulation task, where we employed CLIPort \cite{shridhar2021cliport}. CLIPort is an end-to-end visuomotor policy, trained through imitation learning, that relies on language conditioning and is specifically designed to execute fine-grained manipulation tasks. We made use of a single multitask policy that had been trained to complete a wide range of tabletop tasks, including but not limited to stacking blocks, packing household items, and manipulating deformable objects. As a result of its diverse training, this multitask policy demonstrates an understanding of a wide range of concepts such as colors, common household objects (such as ``a yoshi figure" or ``a soccer cleat"), and directives (including ``push," ``place," and ``pack") \cite{shridhar2021cliport}. 

The intent behind this setup was to evaluate whether our LLM-based anomaly detector can effectively identify when such ``known concepts'' are assembled in a semantically distracting way in the context of a downstream task and compare with existing OOD detection methods. Therefore, to test the policy and our monitor, we define a manipulation task wherein the robot must pick and place blocks into bowls\footnote{This task is adapted from the \texttt{put-blocks-in-bowl} task defined by \cite{shridhar2021cliport}.} and consider three variations of this task:

\begin{figure*}[t]
\begin{center}
\includegraphics[width=\textwidth]{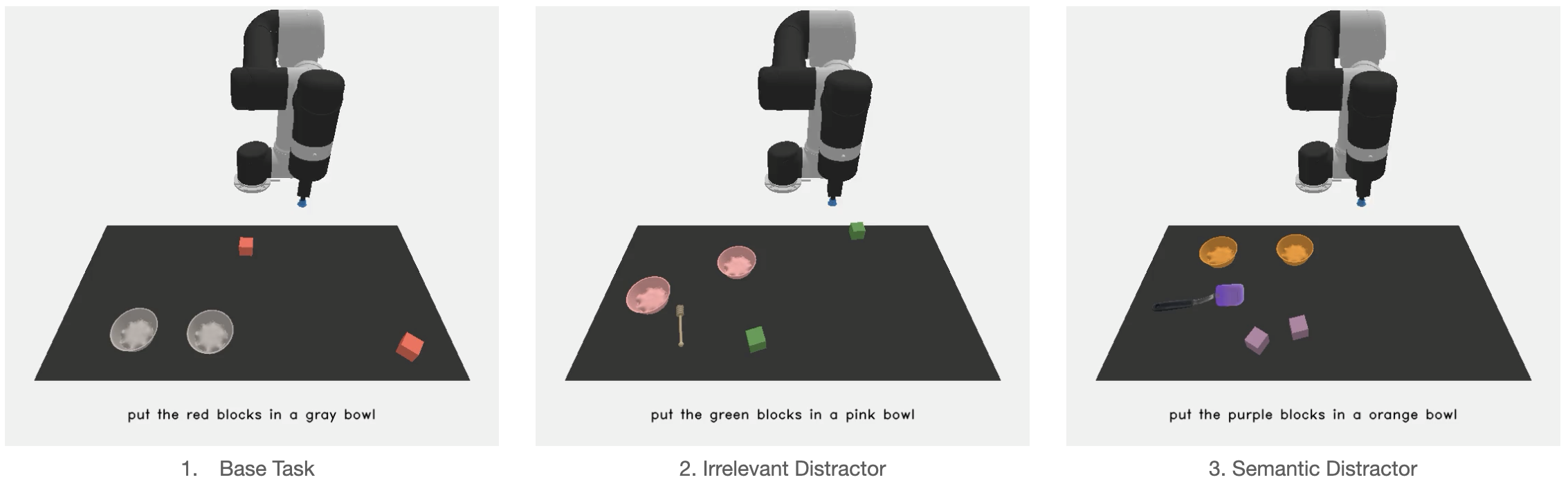}
\end{center}
\caption{An instance of each manipulation task variant: 1. Baseline, 2. Neutral Distractor, 3. Semantic Distractor.}
\label{fig:cliport_examples}
\end{figure*}

\begin{enumerate}
    \item Baseline: Two blocks of a randomly chosen color and two bowls of another randomly chosen color are placed on the workspace. This was one of the tasks used to train the CLIPort policy.
    \item Neutral Distractor: The baseline setup with an additional randomly selected object. The object is visually distinct from the task blocks and bowls in both form and color. An example of one such distractor is the honey dipper seen in Figure \ref{fig:cliport_examples}.
    \item Semantic Distractor: The baseline setup with an additional randomly selected object. The object is meant to visually resemble either the blocks or bowls in some way to challenge the policy. An example of one such distractor is the spatula with a square purple head meant to look like the purple blocks in Figure \ref{fig:cliport_examples}.
\end{enumerate}
We do so because despite its comprehensive multi-task training, the CLIPort policy is still prone to failure, especially on variations of the pick and place task not seen at training time.
Intuitively, we can likely attribute failures when distracting objects are present to contextual reasoning errors induced by spurious correlations, since the multi-task policy has seen many of the distractor objects in other tasks. For example, we may expect the policy to mistake a Butterfinger chocolate in a yellow wrapper for a yellow block even though the policy is aware of the Butterfinger chocolate concept. Therefore, in this experiment we investigate whether the LLM can reason about the differences between neutral and semantically distracting objects.  

We instantiated and simulated 250 episodes of each variant in Ravens \cite{zeng2020transporter}, an environment consisting of a Universal Robot UR5e manipulator attached to the 0.5 m $\times$ 1 m workspace. Neutral and semantic distractors were sampled from the subset of the Google Scanned Objects dataset \cite{downs2022google} used in \cite{shridhar2021cliport} (see Appendix \ref{ref:appendix_distractors}). Examples of each task variant are shown in Figure \ref{fig:cliport_examples}.

\subsubsection{Anomaly Detection Methodology}
For these experiments, we compare our LLM-based anomaly detector with a commonly used OOD detection baseline.

\textbf{LLM Anomaly Detection (Ours):} For these experiments, we prompted the LLM for each neutral or semantic distractor episode to indicate whether the distractor was likely to yield degraded task performance. Prompts were generated by introducing the task specification (e.g., ``put the red blocks in a gray bowl") and a brief visual description of the scene within a template\footnote{In these experiments, we generated these scene descriptions using privileged simulator information. In principle, an object detector could have been used to identify the objects involved in our experiments, however we found that the simulator visuals were not amenable to pretrained detection models.}. The prompt template was designed to elicit a chain-of-thought reasoning style, by querying the LLM to describe the shapes and colors of the blocks, bowls, and distractor and subsequently compare them to identify any possible sources of visual similarity and thus confusion. We refer the reader to the appendix for the full prompt template.

\textbf{Autoencoder Anomaly Detection (Baseline):} As a baseline OOD detection method, we trained a convolutional autoencoder with an L2 reconstruction loss on images from the CLIPort training tasks in Ravens \cite{zeng2020transporter}. The reconstruction error of the autoencoder is a common OOD detection heuristic, because autoencoders learn to reconstruct the training images but fail to generalize on test images that are visually dissimilar from the training data, leading to high reconstruction error OOD \cite{oza2019c2ae, JapkowiczMyersEtAl1995, RuffKauffmanEtAl2021, YangZhouEtAl2021}. Similar to \cite{RichterRoy2017}, we threshold the reconstruction error at the 95\% quantile of the reconstruction losses on the training examples for which the CLIPort policy was successful, yielding a classifier that flags an anomaly when the scene is visually distinct from nominal policy successes.

\begin{figure}
    \centering
    \includegraphics[width=\linewidth]{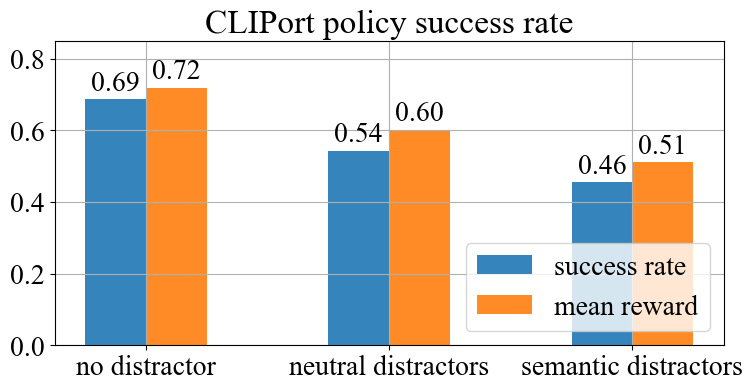}
    \caption{CLIPort policy performance on task variants. An episode is considered successful if all blocks are successfully placed into bowls. The mean reward indicates the average task progress across the set of episodes, with a reward of 0.5 granted for each block that is placed in a bowl.}
    \label{fig:cliport_performance}
\end{figure}
\subsubsection{Results}

\begin{table*}[h!]
\caption{Anomaly detection on the CLIPort pick and place variations with neutral and semantic distractors. The detection rate is the fraction of episodes for which an anomaly was detected out of a total of 250 episodes for each task variation (i.e., the fraction of neutral distractors detected out of 250 episodes with a neutral distractor), for the human baseline, our LLM monitor, and the autoencoder comparison.}
\label{table:llm_human_comparison}
\centering
\begin{tabularx}{\linewidth}{X|X|X|X|}
\hline 
& \multicolumn{3}{c|}{\textbf{Anomaly Detection Rate}} \\
{\bf Task Variant}  & {\bf Human} & {\bf LLM } & {\bf Autoencoder}
\\ \hline 
\textbf{Neutral Distractor} &0 & 0.16  & 0.24\\
\textbf{Semantic Distractor}  & 1.0 &0.86 & 0.51 \\
\hline
\end{tabularx}
\end{table*}
In Figure \ref{fig:cliport_performance}, we compare the policy's performance across the task variants. It is clear that introducing distractors degrades the policy's performance, with the largest drop in performance on average corresponding to the semantic distractor cases. This is reflected both in the number of successful task completions and in the average task progress as measured by the mean reward (a reward of 0.5 is granted for each block that is placed in a bowl). 

Qualitatively, we found that the policy would tend to ``misinterpret" semantic distractors as task-relevant objects. For example, the spatula with a purple head in Figure \ref{fig:cliport_examples} was indeed mistaken for a purple block. However, we also observed a significant proportion of erratic failures across all three experiment variants, such as the manipulator repeatedly reaching for empty areas of the workspace, even in experiments without distractors. The rate of such failures increased in the presence of neutral distractors.



Next, we evaluate the LLM's performance in detecting semantic anomalies. First, we compare the LLM's classification with that of human intuition and show the results in Table \ref{table:llm_human_comparison}. Similar to the human classifications, the LLM flags the majority of semantic distractors and ignores most neutral distractors, thereby supporting the hypothesis that we can use LLMs to reason in alignment with a human's intuition about semantic anomalies. Disagreement tends to arise when the LLM makes decisions on the basis of ambiguous descriptions. For example, when the task involves an orange bowl, the LLM does not detect the presence of ``a light brown fedora" distractor as a potential source of confusion on the basis that ``light brown" and ``orange" are different colors, even though they are actually visually similar.


Secondly, we contrast the LLMs anomaly classifications with the autoencoder baseline, also shown in Table \ref{table:llm_human_comparison}. We see that the autoencoder flags more neutral distractors than the LLM, and only exceeds random guessing on the semantic distractors by a small margin. This indicates that although the episodes with semantic distractors appear too visually similar to in-distribution observations for common OOD detectors to notice, the LLM can still detect semantic anomalies by reasoning about the scene's context.

In addition to evaluating the LLM's performance in detecting semantic anomalies, we also investigate whether its anomaly detection can predict task failures and compare against the autoencoder. Table \ref{table:llm_semantic_prediction} presents confusion matrices for the semantic and neutral distractor experiments.

Our findings indicate that the LLM-based monitor's anomaly detections correspond to a significant number of policy failures when a semantic distractor is present, although the number of false positives is high. However, the monitor understandably struggles to predict failures with only a neutral distractor present. In contrast, the autoencoder monitor has difficulties in both experiment classes. Over the course of these experiments we noticed that a substantial proportion of the policy failures were difficult to trace to a clear cause, since even without distractors present, the failure rate is approximately $30\%$.

Utimately, our evaluations indicate that this anomaly detection framework demonstrates strong alignment with human intuition. However, the unexplainable policy failures confound the failure prediction results motivating future work in grounding the LLM-based monitor with respect to system capabilities (see Section \ref{sec:grounding}).

\begin{table*}[h!]
    \caption{Confusion matrices for fault detection on the CLIPort task variations. Each task variation consists of 250 episodes, for which we report the total number episodes on which the policy succeeded or failed (task success/failure) vs. the number of anomalies detected or missed. We report the confusion matrix for both our LLM detector and the autoencoder baseline.}
    \centering
    \begin{tabularx}{\textwidth}{XX|XX|XX|}
        \hline
        & & \multicolumn{2}{c|}{\textbf{LLM}} & \multicolumn{2}{c|}{\textbf{Autoencoder}}\\
        & & Anomalies Detected & Anomalies Missed & Anomalies Detected & Anomalies Missed
        \\ \hline 
        \multirow{2}{*}{\textbf{Semantic}} & \textbf{Task Success} &103 &11 & 57 & 57\\
        & \textbf{Task Failure} &113 &23 & 70 & 66 \\
        \hline
        \multirow{2}{*}{\textbf{Neutral}} & \textbf{Task Success} &22 &114 &  27 & 109\\
        & \textbf{Task Failure} &17 &97 & 33&  81\\
        \hline
    \end{tabularx}

\label{table:llm_semantic_prediction}
\end{table*}


%% file: tex/5-discussion.tex
\section{Discussion}\label{sec:discussion}
The experimental results demonstrate the potential for LLMs to advance in-context and task-relevant anomaly detection capabilities. The autonomous driving experiments illustrate that the LLM-based detection methodology exhibits the ability to identify semantic anomalies that could arise in complex and diverse scenarios. Furthermore, the manipulation experiments highlight an LLM's capacity to disentangle the subtle semantic distinctions that give rise to anomalous behavior, which classical OOD methods are not equipped to handle. In this section, we will present a series of illustrative examples to elucidate the strengths and limitations of our proposed methodology, and then proceed to discuss potential remedies for these limitations.

\begin{figure*}[h]
\centering
\subfigure[Encountering a truck carrying a traffic light.]{\includegraphics[width=0.45\linewidth]{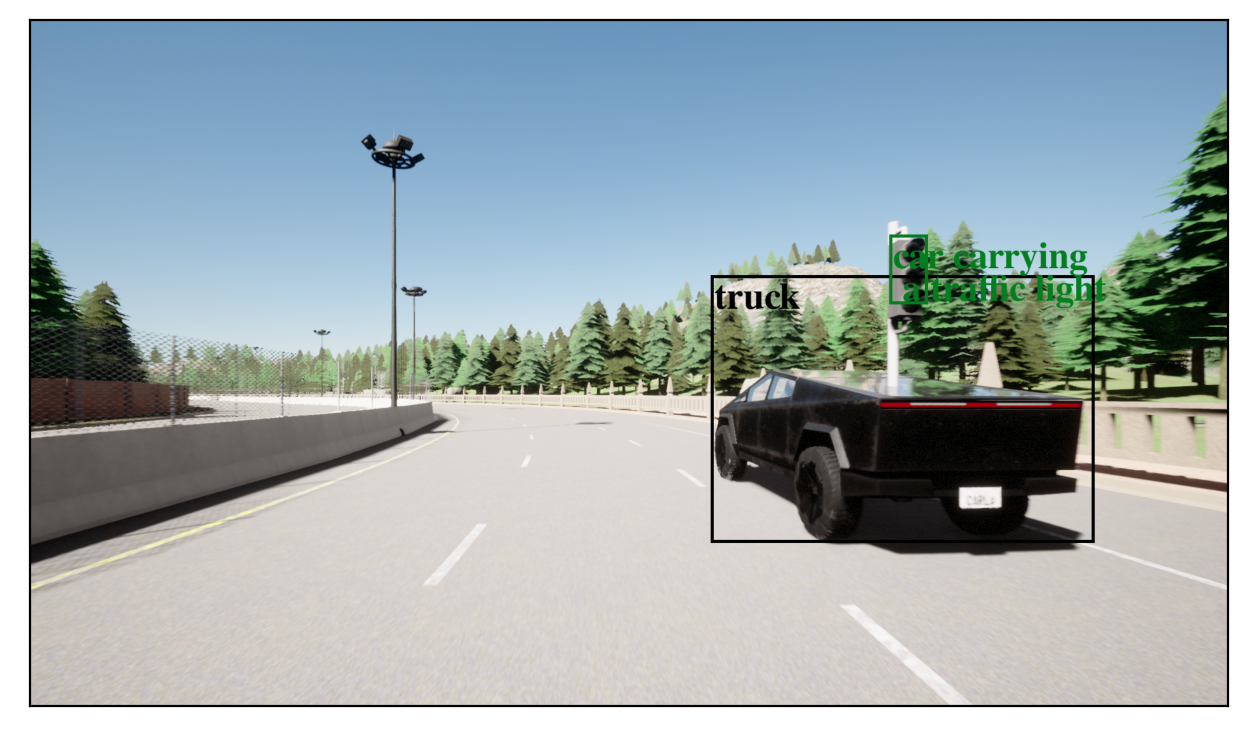}}
\subfigure[A billboard with realistic stop sign imagery.]{\includegraphics[width=0.45\linewidth]{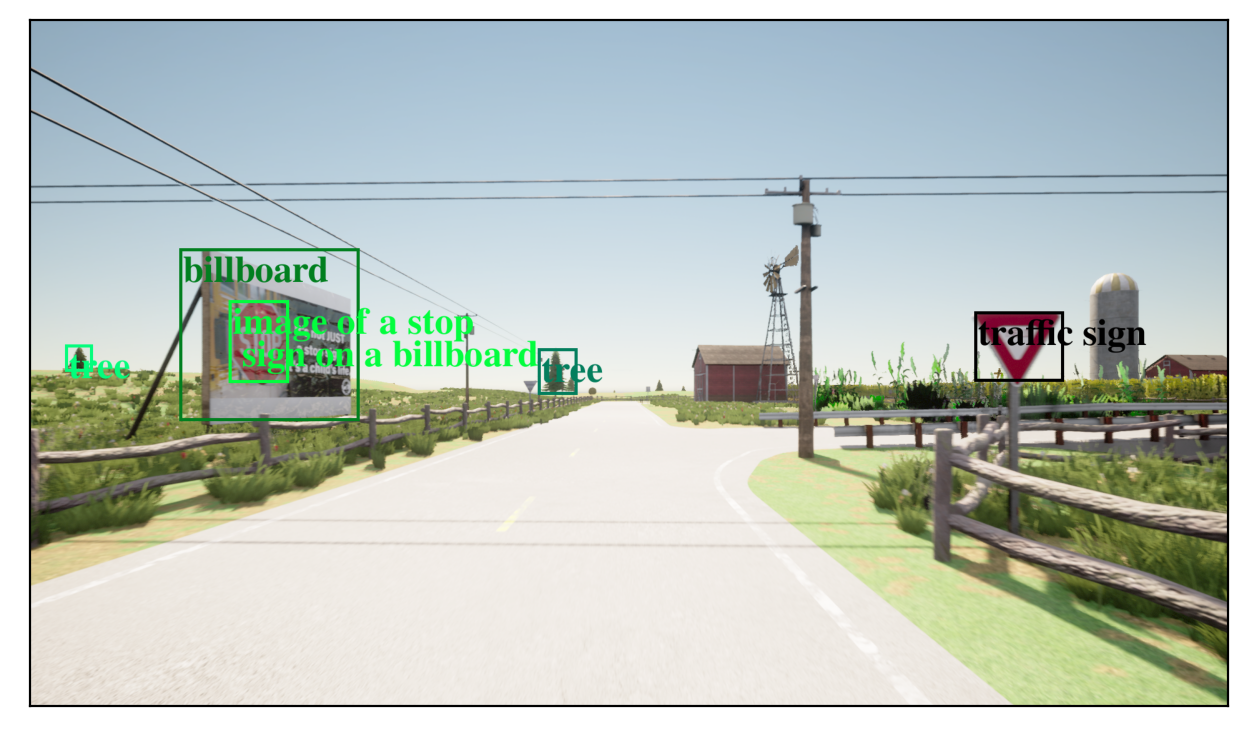}}
\subfigure[An normal traffic light intersection.]{\includegraphics[width=0.45\linewidth]{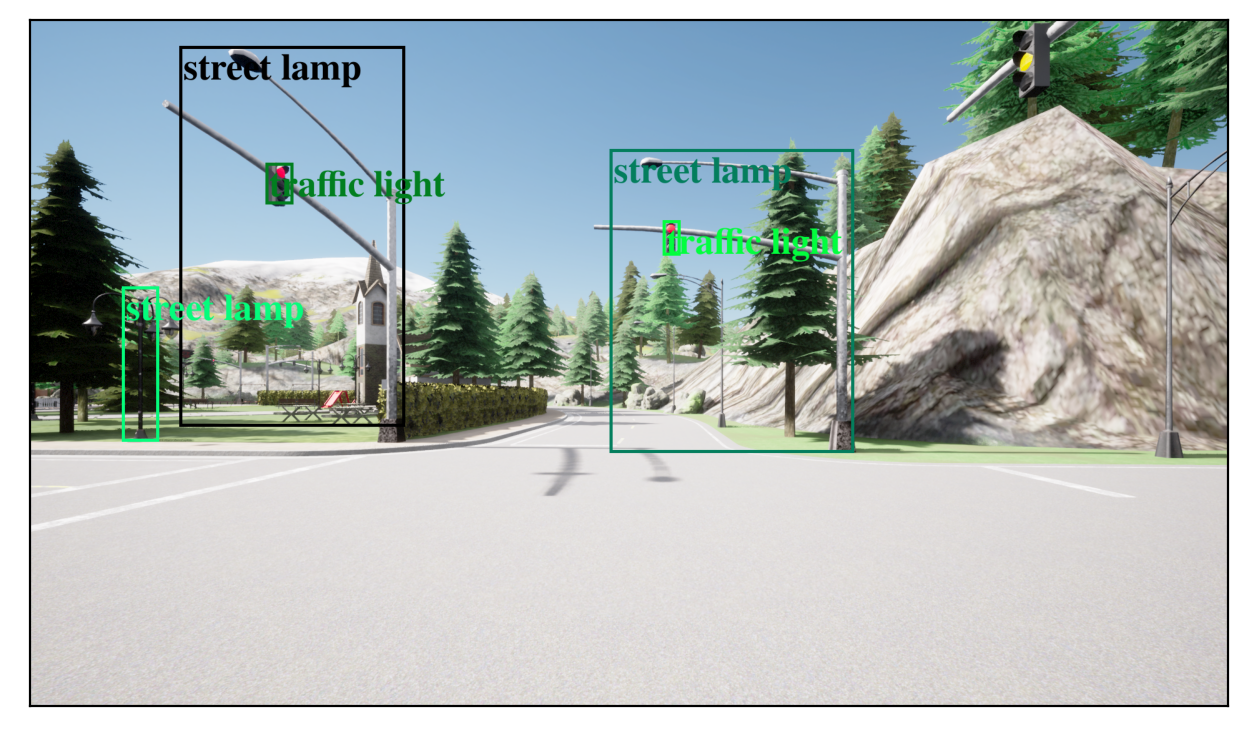}}
\subfigure[A normal stop sign intersection.]{\includegraphics[width=0.45\linewidth]{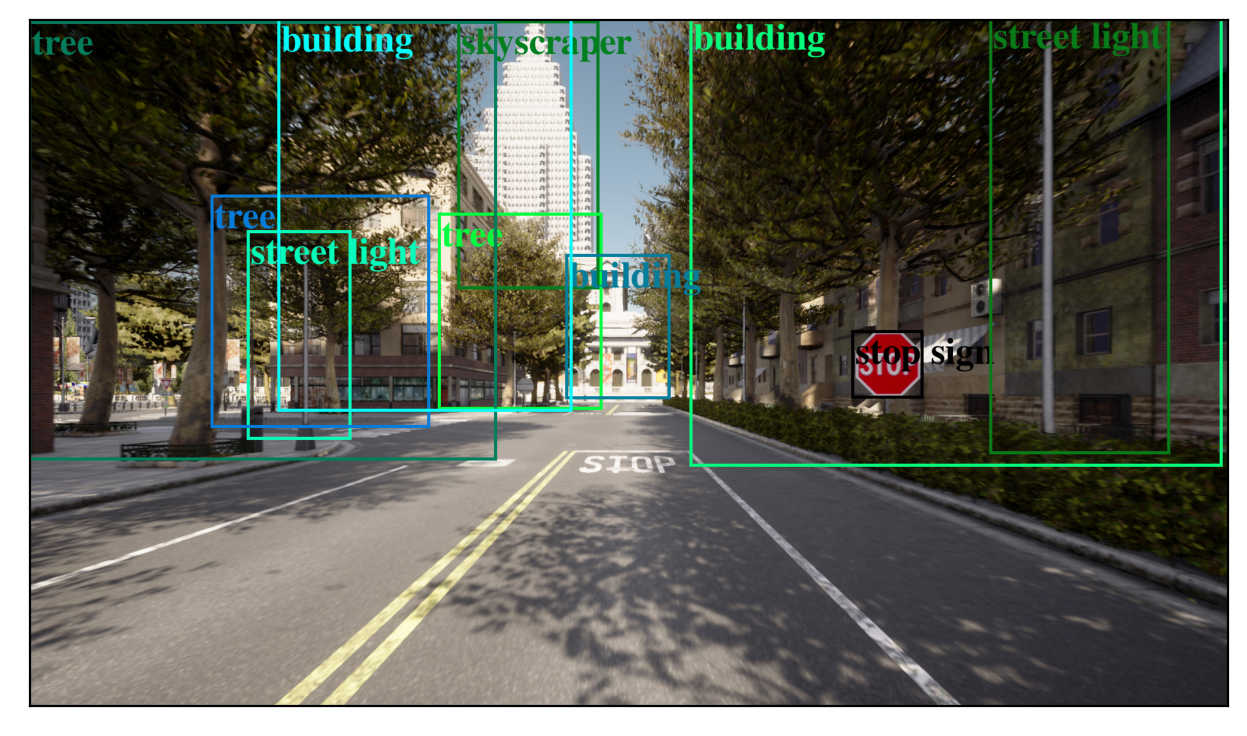}}
\subfigure[An observation yielding a false positive anomaly warning. The LLM incorrectly indicates that the ``building by the road" could ``block the vehicle's view of the road or other objects."]{\includegraphics[width=0.45\linewidth]{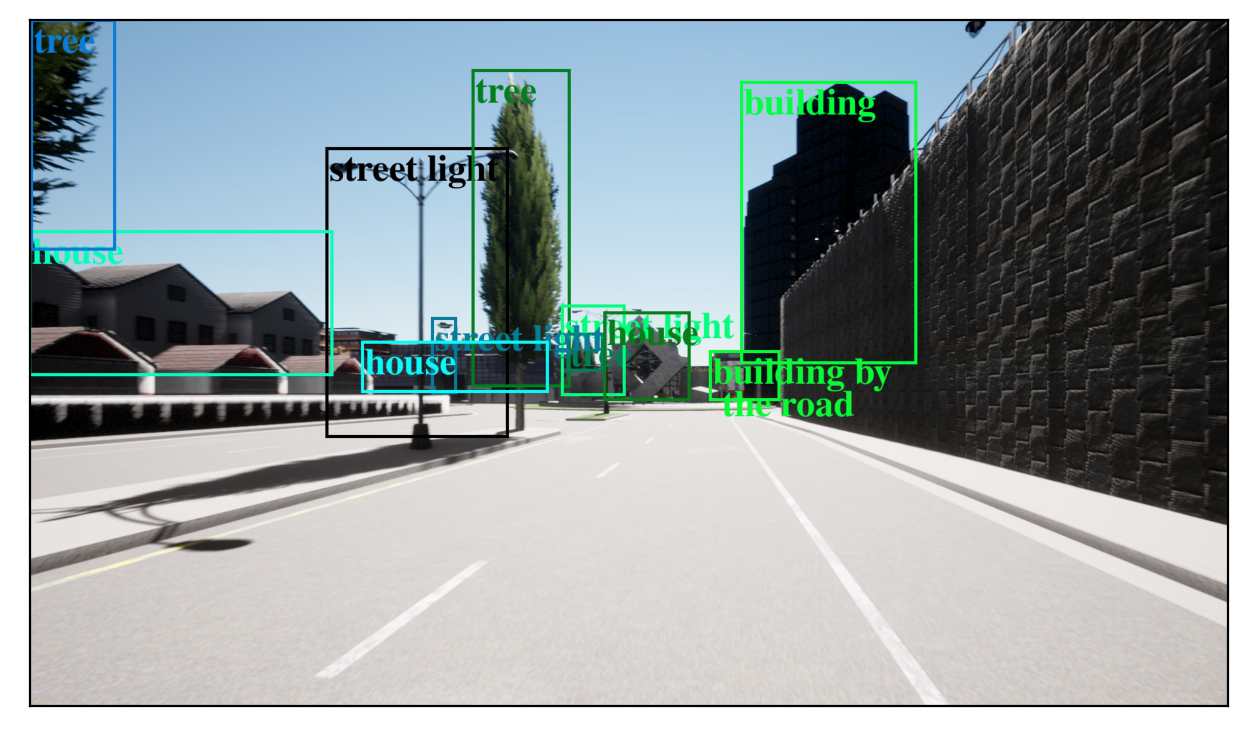}}
\subfigure[An observation corresponding to a missed anomaly. The LLM incorrectly indicates that the ``image of a stop sign on a billboard" is normal since the ``autonomous vehicle should be able to recognize and obey stop signs."]{\includegraphics[width=0.45\linewidth]{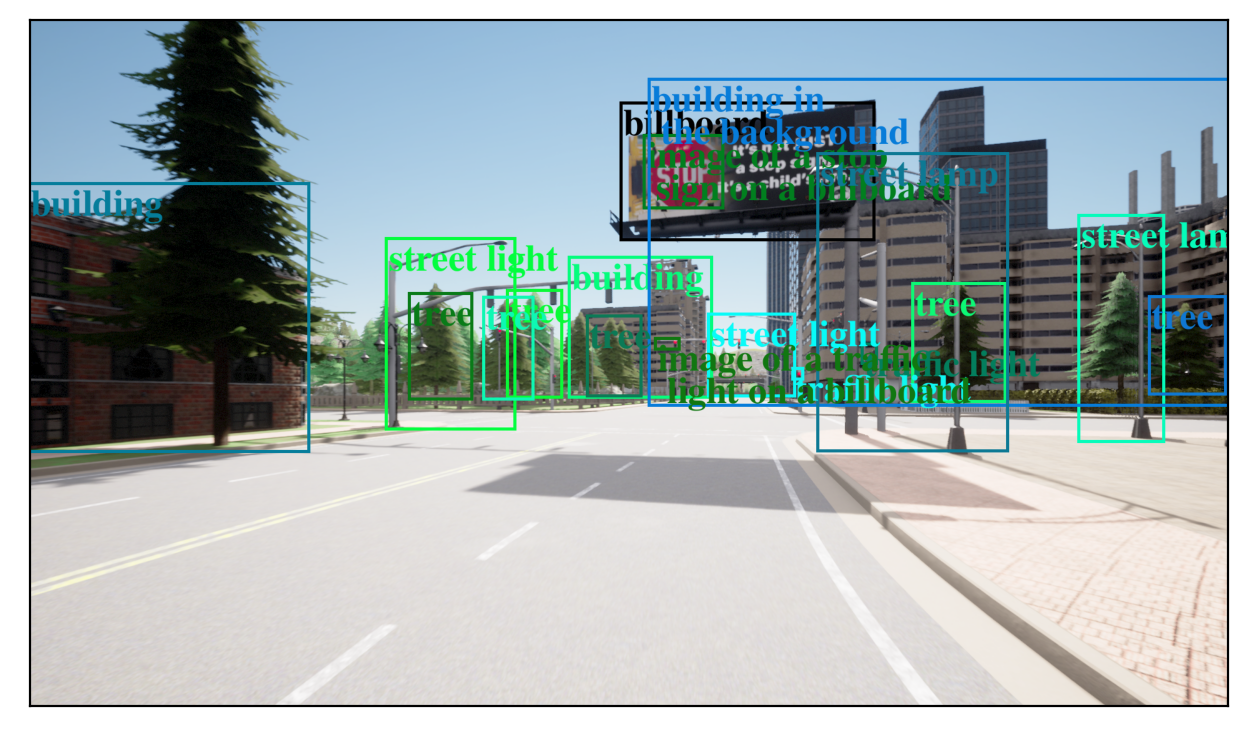}}
\caption{Illustrative observations from various simulated autonomous driving scenarios. Images (a) and (b) correspond to images that the LLM-based monitor correctly identifies as semantic anomalies. Images (c) and (d) depict common scenes that the monitor correctly identifies as normal. Images (e) and (f) demonstrate mistakes made by the monitor.}
\label{fig:carla_discussion}
\end{figure*}


\subsection{Qualitative Analysis}

In order to provide a more detailed insight into the performance of our methodology, we turn our attention to a series of illustrative examples. In particular, we present six examples from our autonomous vehicle experiments in Figure \ref{fig:carla_discussion}. 

Images (a) and (b) showcase the effectiveness of the LLM-based approach in detecting anomalies. These cases respectively correspond to reconstructions of the aforementioned Tesla failure cases; the first simulates the vehicle's encounter with a truck transporting an inactive traffic light and the second displays the vehicle's observation of a billboard with realistic stop sign imagery. Following these, images (c) and (d) depict unmodified, common traffic light and stop sign intersections, which the LLM correctly identifies as non-anomalous.

Notably, these cases demonstrate the LLM's capabilities to perform nuanced contextual reasoning. In the case of the stop sign on the billboard, the LLM recognizes that it is an image on a billboard and is not functional. Similarly, when confronted with the traffic light on the truck, the LLM can identify that it is not in its usual position but rather is being transported by the truck. In contrast, the LLM considers the observation of a simple stop sign or traffic light to be typical and well within the capabilities of an autonomous vehicle.

The final pair of images in Figure \ref{fig:carla_discussion} highlight two examples that demonstrate the types of reasoning errors made by the LLM. Image (e) features a false anomaly alert due to an innocuous sighting of ``a building by the road." Curiously, such detections in other observations do not trigger a warning and are appropriately deemed commonplace scenery. Image (f) similarly depicts an error in LLM reasoning, albeit due to a missed anomaly this time. Here, the vehicle perceives a stop sign on a billboard, which the LLM deems a normal observation since "the vehicle should be able to recognize and obey stop signs." Interestingly, for the aforementioned failure cases we find that the LLM's responses are sensitive to the scene description order. Manually rearranging the order of scene elements in the prompts for these two examples was observed to invert the the anomaly classifications, suggesting the need to calibrate the monitor's outputs.

More generally, we qualitatively observe that the LLM's analysis tends to exploit the provided few-shot examples quite heavily as opposed to truly analyzing the scene description. We note that the provided justifications for the anomaly characterizations often closely reflect or overfit to the reasoning provided in the examples, with limited novelty. Zero-shot prompting was found to leverage the LLM's latent knowledge most effectively, yielding the most creative responses at the cost of task-relevance and reliability, limiting their utility for anomaly detection. As of yet, this novelty-relevance trade off is a matter of prompt tuning, but upcoming foundation models are suggested to feature improved creativity and reasoning capabilities \cite{bubeck2023sparks, openai2023gpt4}. We further discuss constraining the LLM's responses with respect to task-relevant and system-capabilities in Section \ref{sec:grounding}.

In light of the current limitations and ongoing foundation model advances, we present a discussion on possible avenues for future development of semantic anomaly detection in the following section.

\subsection{Research Outlook}

\subsubsection{Multimodal Context}

Accurate and comprehensive scene descriptions are crucial for effective semantic anomaly detection using LLMs. An LLM's ability to distinguish normal from anomalous observations often relies on subtle qualifications of scene descriptions. For instance, distinguishing between a ``truck carrying a traffic light" and a common ``traffic light" results in the former being considered an anomaly and the latter being a commonplace observation.

However, natural language, while a powerful and flexible means of conveying information, has a fundamental limitation in producing ambiguous or underspecified descriptions of scenes and objects. Consider the observation of a ``house on the sidewalk," which the LLM wrongly flags as an anomaly due to the lack of specificity and context in the object description. Instead of interpreting this description to indicate a house located near the sidewalk, the LLM understands it to mean a house physically placed on the sidewalk, obstructing the road. Crucially, the lack of specificity and limited context afforded by the object description allow the LLM to justify this misclassification.

To address these issues, it is essential to move beyond purely natural language prompting. Multimodal models such as GPT-4 \cite{openai2023gpt4} and Flamingo \cite{alayracflamingo} can incorporate both visual and textual inputs to provide a more direct means of conveying the robot's observations. Directly incorporating images into the prompt can better preserve context, overcoming the limitations associated with natural language descriptions and minimizing information losses incurred through the visual parsing process.

\subsubsection{Grounding System Capabilities}\label{sec:grounding}

Although LLMs are be trained on general datasets which provide a broad basis for contextual reasoning, a target system possesses a specific set of skills. In order to effectively integrate an LLM-based anomaly detector within an autonomy stack, it is necessary to ``inform" the LLM of these system-specific capabilities and limitations. For example, an autonomous vehicle programmed for urban environments may have different failure modes and anomalous responses than one designed to operate in rural areas. 

To an extent, this is achieved through few-shot and chain-of-thought prompting, as the provided examples and logical structure ``prime" the LLM as to how to characterize anomalies. However, this is an indirect strategy that only conveys a system's competencies implicitly and incompletely. Furthermore, this requires manual prompt engineering on a per-system basis, is subject to the designers' biases and oversights, and does not account for account for system improvements or updates over time. More comprehensive solutions merit consideration, such as fine-tuning the LLM on system-specific data or more tightly integrating the LLM within the autonomy stack to jointly reason over observations, states and actions.




\subsubsection{Complementary Anomaly Detection and Mitigation}

Our LLM-based anomaly detection methodology offers a powerful tool for monitoring robot observations at a contextual level. However, data-driven systems may sometimes fail unexpectedly or inexplicably, as evidenced by our CLIPort experiments. In light of this, we recommend conducting a more comprehensive study to investigate how our method can complement other OOD detection methods that are better suited to identifying and explaining such failure modes. This would enable us to develop a more complete and robust monitoring suite that can provide deeper insights into system performance and potential issues.

Beyond anomaly detection, fault mitigation and recovery are important related considerations. Although these problems are beyond the scope of this work, an LLM-based approach to anomaly detection allows for the formulation of structured queries to either initiate system recovery or request supervisor assistance, which could facilitate the process of devising contingency strategies or simplifying the required human intervention required. In fact, recent works have demonstrated that LLMs can generate feedback to propose improvements or corrections for control tasks \cite{srivastava2023generating}, produce an executable plan to rectify failures based on recent experiences \cite{liu2023reflect}, or simply ask a human for help when needed \cite{ren2023robots}.


%% file: tex/6-conclusion.tex
\section{Conclusion}\label{sec:conclusion}

As modern robotic systems continue to evolve and become more sophisticated, the possibility of encountering challenging corner cases that may degrade system performance or pose safety hazards persists. The semantic anomalies that this study aims to address represent a particularly difficult subset of these cases, requiring a level of insight and reasoning capabilities akin to those of humans. Given their extensive training data, largely derived from human experience, LLMs possess the contextual knowledge and demonstrate strong emergent reasoning capabilities, making them compelling tools for semantic anomaly detection.

The proposed framework leverages LLMs' reasoning abilities to detect semantic anomalies. Our experiments have demonstrated that the LLM-based monitor matches human intuition in both fully end-to-end policies and classical autonomy stacks that utilize learned perception. We found that semantic anomalies do not always correspond to semantically-explainable failures, particularly for end-to-end policies which can behave erratically. However, LLM-based monitoring lends itself to multiple avenues of further developement, such as by 1) more tightly coupling robot perception to LLM prompts, 2) grounding LLM outputs with respect to system capabilities, and 3) investigating the complementarity of the LLM-based monitor with other OOD/anomaly monitors.



%% file: tex/declarations.tex
\section*{Declarations}
\subsection{LLM Usage Statement}
LLMs were exclusively used for the experimentation and evaluation of our proposed methodology. OpenAI's \texttt{text-davinci-003} model was used for all experiments. LLMs were not used in the composition of this manuscript.

\subsection{Funding}
The NASA University Leadership initiative (grant \#80NSSC20M0163) and Toyota Research Institute provided funds to support this work. Amine Elhafsi is supported by a NASA NSTGRO fellowship (grant \#80NSSC19K1143). This article solely reflects the opinions and conclusions of its authors and not any NASA, TRI, or Toyota entity.

\subsection{Competing Interests}
Not applicable.

\subsection{Compliance with Ethical Standards}
Not applicable.

\subsection{Consent to Participate}
Not applicable.

\subsection{Consent for Publication}
The authors unanimously endorsed the content and provided explicit consent for submission. They also ensured that consent was obtained from the responsible authorities at the institute(s)/organization(s) where the research was conducted.

\subsection{Availability of Data and Materials}
Relevant documentation, data and/or code is readily available to verify the validity of the results presented upon request.

\subsection{Code Availability}
The authors plan to open source their code.

\subsection{Author Contributions}
\textbf{Amine Elhafsi} initiated the project, developed the methodology, performed prompt tuning, and implemented and conducted the experiments. \textbf{Rohan Sinha} prepared the structure for the CARLA autonomous vehicle stack, conducted autonomous vehicle experiments, computed autoencoder OOD detector baseline metrics, processed experimental results, and performed data analysis. \textbf{Christopher Agia} implemented the autoencoder OOD detector baseline for the learned policy experiments. \textbf{Edward Schmerling} implemented the autonomous vehicle traffic light classification, performed data analysis, and advised the project. \textbf{Issa A. D. Nesnas} advised the project. \textbf{Marco Pavone} was the primary advisor for the project. The manuscript was jointly written by Amine, Rohan and Edward. All authors reviewed and revised the manuscript.

%% file: tex/appendices.tex









\section{Additional Details: Reasoning-Based Policy}\label{sec:carla_prompt}

The following template was designed to prompt an analysis of the autonomous vehicle's scene observations. Placeholders are indicated by the braces and are substituted for the relevant information at each query.

\vspace{5mm}

\begin{mdframed}[backgroundcolor=light-gray, roundcorner=10pt,leftmargin=1, rightmargin=1, innerleftmargin=15, innertopmargin=15,innerbottommargin=15, outerlinewidth=1, linecolor=light-gray]
\begin{lstlisting}[linewidth=\columnwidth,breaklines=true]
I am the fault monitor for a vision-based autonomous vehicle. My job is to analyze the vehicle's observations and identify anything that could cause the vehicle to take actions that are unsafe, unpredictable or violate traffic rules. For each object that the vehicle observes, I will reason about whether the object constitutes a normal observation or an anomaly. Normal observations do not detrimentally affect the vehicle's performance, whereas anomalies might. Finally, I will classify whether the overall scene is normal or abnormal. For example,
"The vehicle is driving on the road and observes:
-a cyclist on the sidewalk
-a car on the road
-a pedestrian carrying a bright green balloon
Cyclist on the sidewalk:
1. Is this common to see while driving? Yes, cyclists can often be seen riding on the road or occasionally on sidewalks.
2. Can this influence the vehicle's behavior? No, they are on the sidewalk and not on the road.
3. Can the vehicle drive safely in its presence? Yes, cyclists are commonly seen on the road and the vehicle should be able to drive safely in their presence.
4. Can this cause the vehicle to make unpredictable or unsafe maneuvers? No, they are on the sidewalk and not on the road.
Classification: Normal.
Car on the road:
1. Is this common to see while driving? Yes, cars are common to see while driving.
2. Can this influence the vehicle's behavior? Yes, the autonomous vehicle must respect other vehicles on the road, avoid collisions and obey the rules of the road.
3. Can the vehicle drive safely in its presence? Yes, cars are commonly seen on the road and the autonomous vehicle should be able to drive safely in their presence.
4. Can this cause the vehicle to make unpredictable or unsafe maneuvers? No, autonomous vehicles are programmed to appropriately drive and interact with other cars on the road .
Classification: Normal.
Pedestrian carrying a bright green balloon:
1. Is this common to see while driving? Pedestrians are commonly seen on the sidewalk or crossing at intersections. They may possess arbitrary objects and a balloon is a reasonable object to carry.
2.  Can this influence the vehicle's behavior? Yes, the autonomous vehicle may mistake the green balloon for a green traffic light signal, which could deceive it into driving forward when it should otherwise be stopped (e.g., at a red light).
3. Can the vehicle drive safely in its presence? No, this could deceive the vehicle into interpreting the green balloon as a legal traffic signal.
4. Can this cause the vehicle to make unpredictable or unsafe maneuvers? Yes, this could deceive the autonomous vehicle into driving forward when it should otherwise be stopped (e.g., at a red light)
Classification: Anomaly.
Overall Scenario Classification: Anomaly.
"
I am driving on the road and I see:
{scene_description}
\end{lstlisting}
\end{mdframed}

\clearpage
\section{Additional Experimental Details: Learned Policy}\label{sec:appendix_learned_policy} 

\subsection{Prompt Template}

The following prompt was designed to elicit a comparison of the distractor objects and the blocks and bowls from the LLM. Placeholders are indicated by the braces and are substituted for the relevant information at each query.

\vspace{5mm}

\begin{mdframed}[backgroundcolor=light-gray, roundcorner=10pt,leftmargin=1, rightmargin=1, innerleftmargin=15, innertopmargin=15,innerbottommargin=15, outerlinewidth=1, linecolor=light-gray]
\begin{lstlisting}[linewidth=\columnwidth,breaklines=true]
You are a robot's fault monitor responsible for detecting situations that could confuse the robot and adversely affect its performance. The robot is a manipulator attached to a table and its task is to pick up blocks of a specified color and place them into bowls of a specified color. However, the robot's vision is prone to making mistakes and may misinterpret an object as a block or a bowl if that object has a similar appearance. Your job is to analyze what the robot sees and identify any objects that may cause it to make a mistake in performing its task. You should flag any objects that have any similarity with the blocks or bowls just in case.

The robot's task is to put the {block_color} blocks in a {bowl_color} bowl.

On the table, the robot sees
{scene_objects}

Analyze each object. First, describe the appearance of each object focusing on the shape and color of each constituent component. Then, indicate whether an object or part of an object may resemble the {block_color} block or {bowl_color} bowl in any way, and whether the robot is likely to misidentify the object.

For each object follow the template:
<Object>:
Object Description:
Object Similarity with Block Shape (explain):
Object Similarity with Block Color (explain):
Misidentifiable as Block (yes or no):
Object Similarity with Bowl Shape (explain):
Object Similarity with Bowl Color (explain):
Misidentifiable as Bowl (yes or no):

After all of the objects have been analyzed, indicate whether there any of the objects can cause the robot to make a mistake. Indicate this with the template:
Misidentifiable Objects Present (yes or no):
\end{lstlisting}
\end{mdframed}

\vspace{5mm}

We chose to abstain from using few-shot prompting for this set of experiments. We noted that the diversity exhibited by the common household object classes used as distractors (as compared to driving objects classes, such as traffic lights and signals exhibit some degree of standardization features) necessitated some degree of zero-shot reasoning by the LLM. This zero-shot prompting strategy encouraged the LLM to leverage its inherent knowledge of common objects more effectively. In contrast, when few-shot prompted, we found that the responses tended to overfit to the provided examples, negatively impacting the LLM's function as a monitor.

\clearpage
\onecolumn
\subsection{Semantic and Neutral Distractors}\label{ref:appendix_distractors}
\needspace{3\baselineskip} 
\begin{longtable}{|>{\centering\arraybackslash}m{0.3\linewidth}|>{\centering\arraybackslash}m{0.55\linewidth}|}
  \caption{Distractors used in the learned policy experiments. We include an image of each distractor alongside its usage in the experiment. Note that semantic distractors were also used as neutral distractors when the task involved blocks or bowls with colors dissimilar to those of the object for additional variety.} \\
  \hline
  \textbf{Image} & \textbf{Description} \\
  \hline
  \endfirsthead
  \multicolumn{2}{c}{{\tablename\ \thetable{} -- continued from previous page}} \\
  \hline
  \textbf{Image} & \textbf{Description} \\
  \hline
  \endhead
  \hline
  \multicolumn{2}{r}{{Continued on next page}} \\
  \endfoot
  \hline
  \endlastfoot
  
  \includegraphics[width=3.5cm]{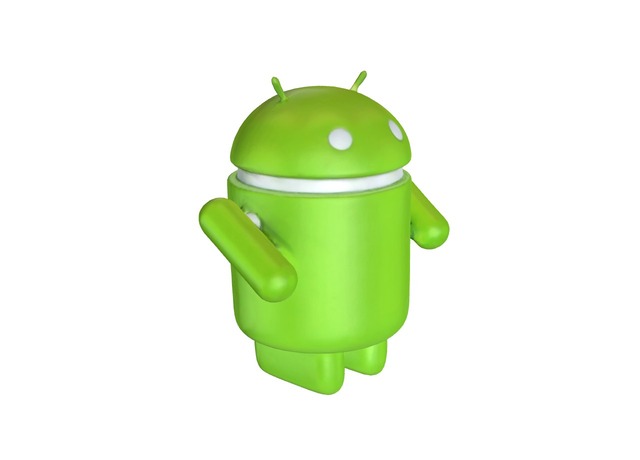} & Android Toy: Semantic distractor for green block. \\
  \hline
  \includegraphics[width=3.5cm]{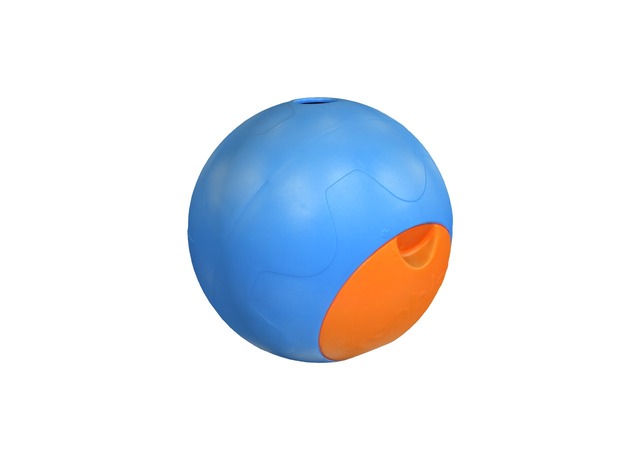} & Ball Puzzle: Semantic distractor for blue bowl. \\
  \hline
  \includegraphics[width=3.5cm]{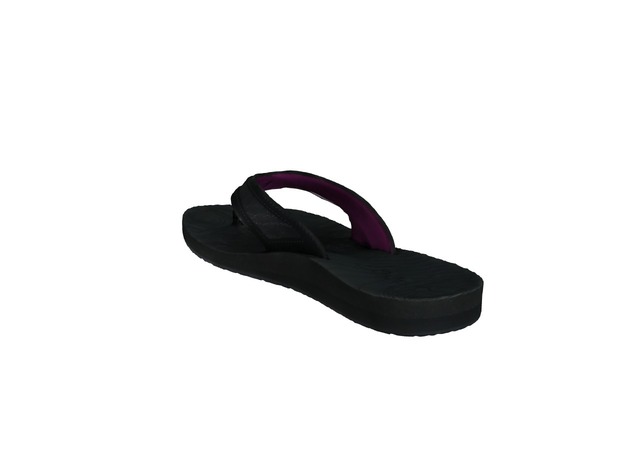} & Black Sandal: Neutral distractor. \\
  \hline
  \includegraphics[width=3.5cm]{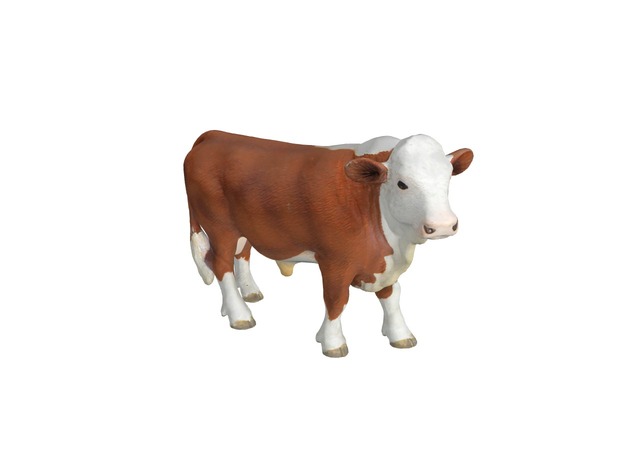} & Bull Figure: Semantic distractor for brown block. \\
  \hline
  \includegraphics[width=3.5cm]{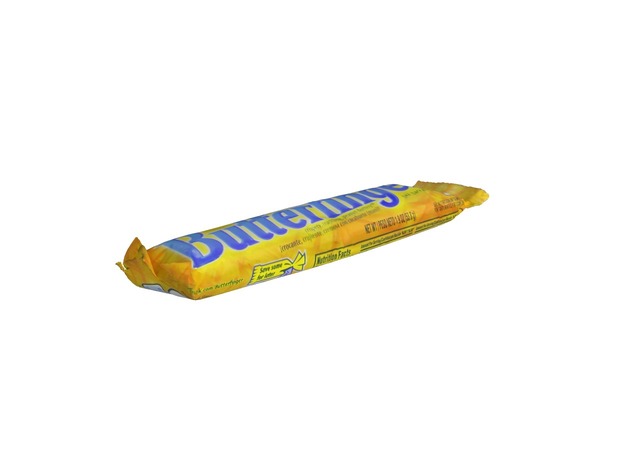} & Butterfinger Chocolate: Semantic distractor for yellow and orange blocks. \\
  \hline
  \includegraphics[width=3.5cm]{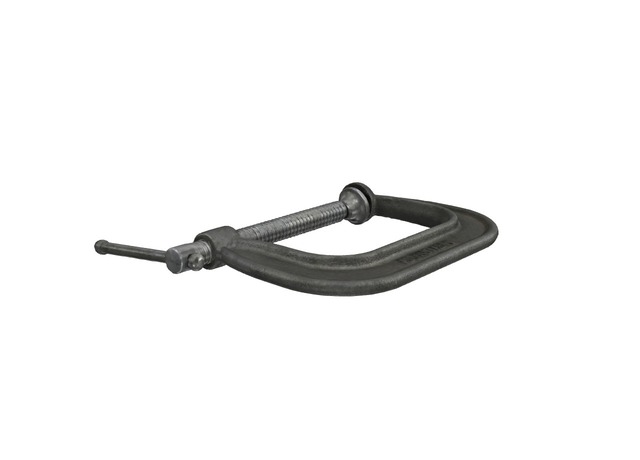} & C-Clamp: Neutral distractor. \\
  \hline
  \includegraphics[width=3.5cm]{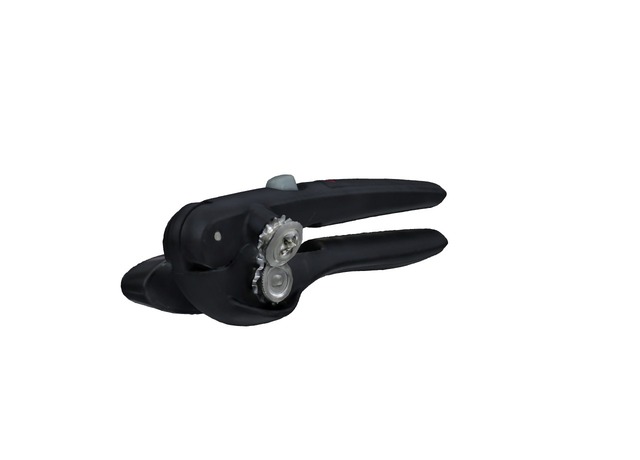} & Can Opener: Neutral distractor. \\
  \hline
  \includegraphics[width=3.5cm]{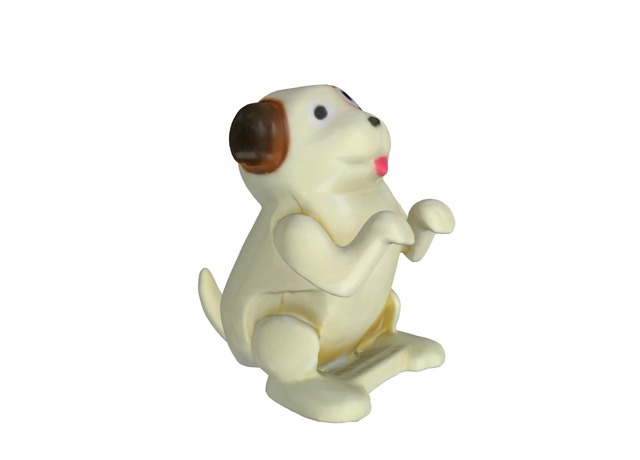} & Dog Toy Statue: Semantic distractor for white block. \\
  \hline
  \includegraphics[width=3.5cm]{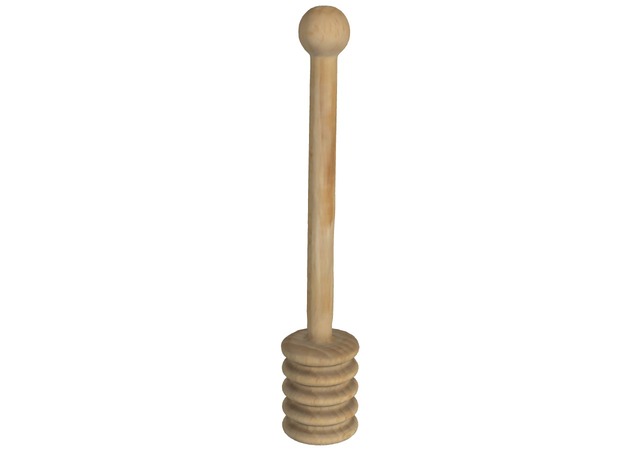} & Honey Dipper: Neutral distractor. \\
  \hline
  \includegraphics[width=3.5cm]{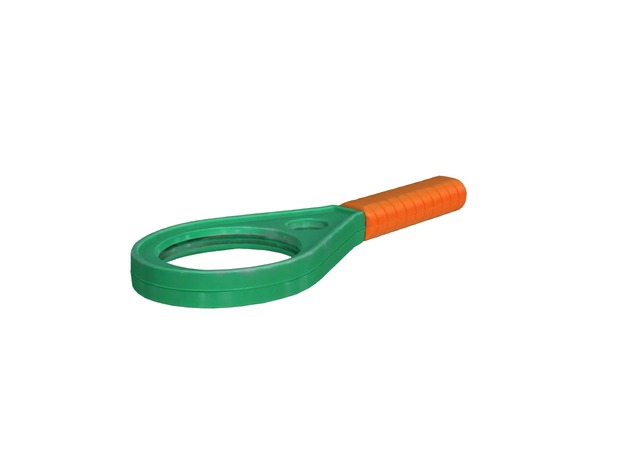} & Magnifying Glass with Green Ring: Semantic distractor for green bowl. \\
  \hline
  \includegraphics[width=3.5cm]{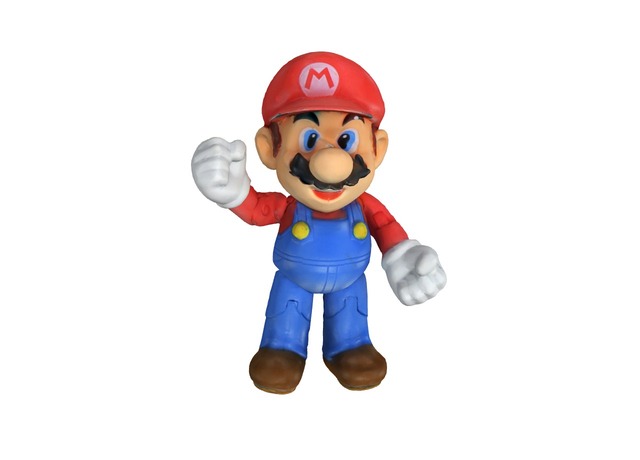} & Mario Figure: Neutral distractor. \\
  \hline
  \includegraphics[width=3.5cm]{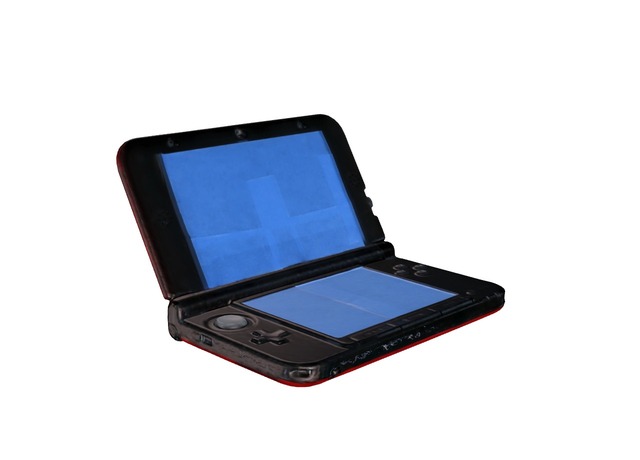} & Nintendo 3DS: Neutral distractor. \\
  \hline
  \includegraphics[width=3.5cm]{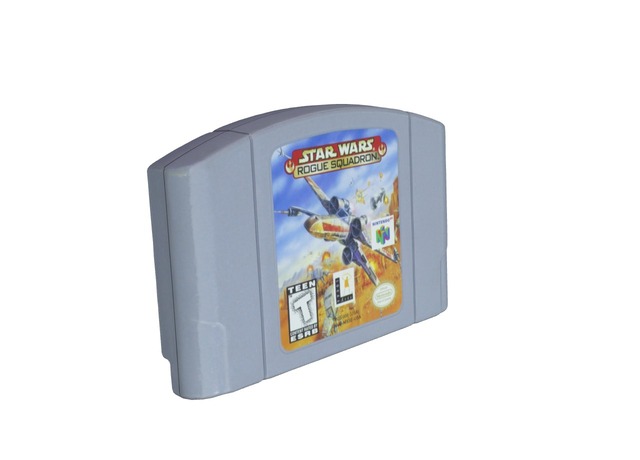} & Nintendo Cartridge: Semantic distractor for gray block. \\
  \hline
  \includegraphics[width=3.5cm]{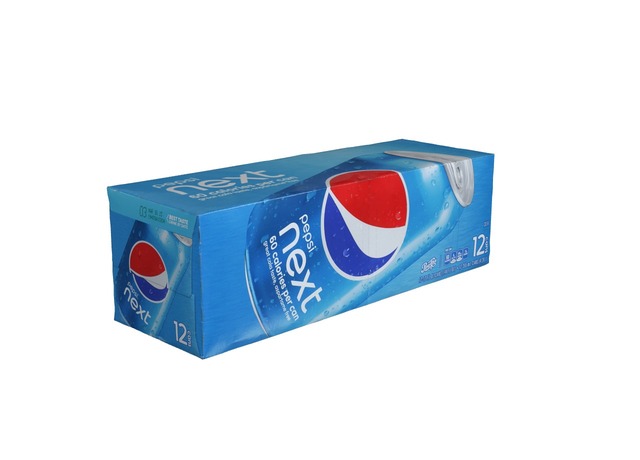} & Pepsi Next Box: Semantic distractor for blue block. \\
  \hline
  \includegraphics[width=3.5cm]{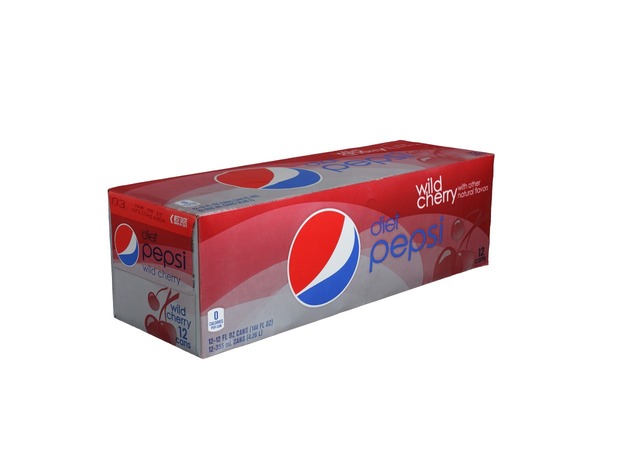} & Pepsi Wild Cherry Box: Semantic distractor for red block. \\
  \hline
  \includegraphics[width=3.5cm]{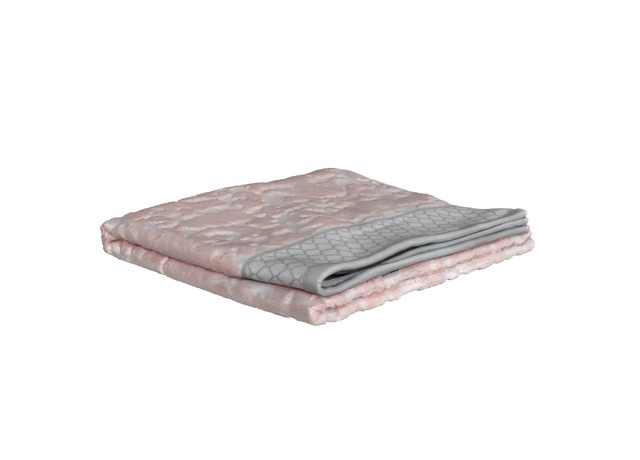} & Pink Towel: Semantic distractor for pink and white blocks. \\
  \hline
  \includegraphics[width=3.5cm]{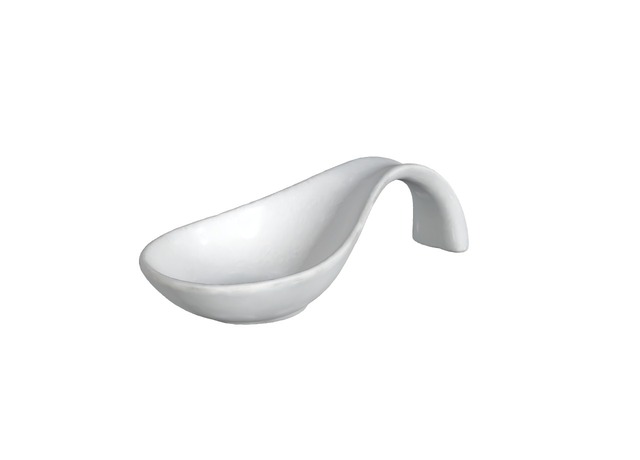} & Porcelain Spoon: Semantic distractor for white bowl. \\
  \hline
  \includegraphics[width=3.5cm]{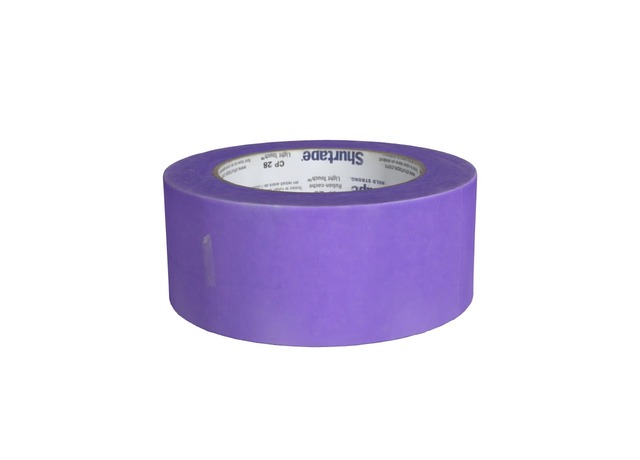} & Purple Tape: Semantic distractor for purple bowl. \\
  \hline
  \includegraphics[width=3.5cm]{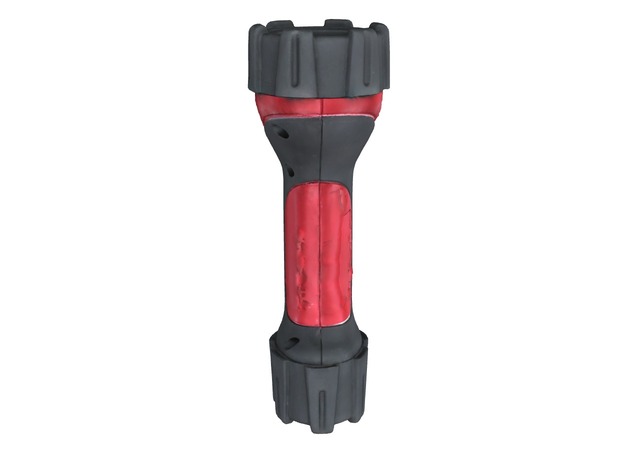} & Flashlight: Neutral Distractor. \\
  \hline
  \includegraphics[width=3.5cm]{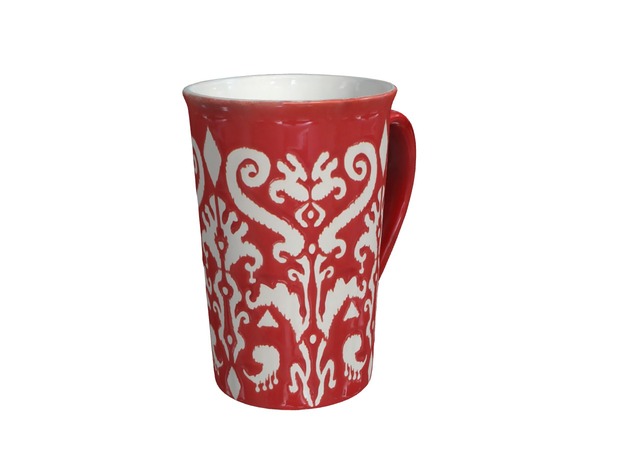} & Red Cup: Semantic distractor for red bowl. \\
  \hline
  \includegraphics[width=3.5cm]{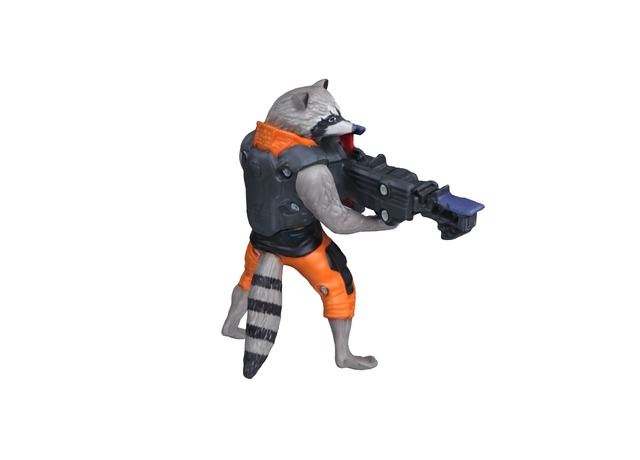} & Rocket Raccoon Figure: Neutral distractor. \\
  \hline
  \includegraphics[width=3.5cm]{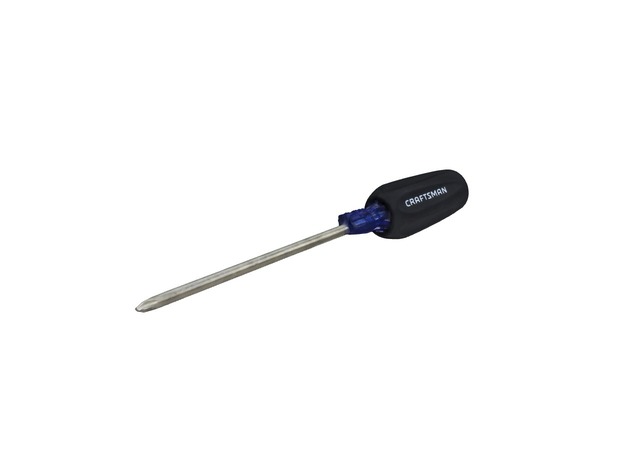} & Screw Driver: Neutral distractor. \\
  \hline
  \includegraphics[width=3.5cm]{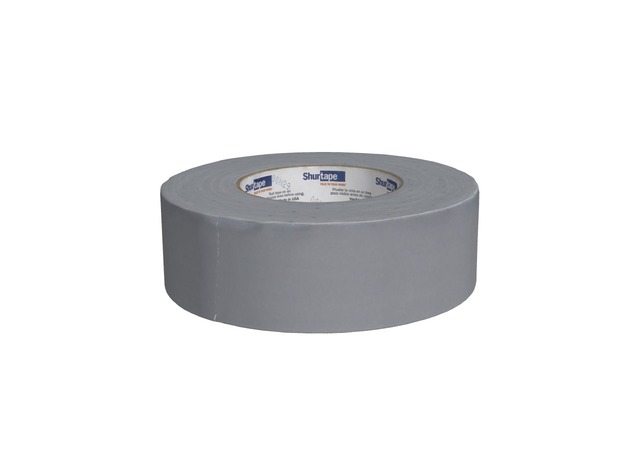} & Silver Tape: Semantic distractor for gray bowl. \\
  \hline
  \includegraphics[width=3.5cm]{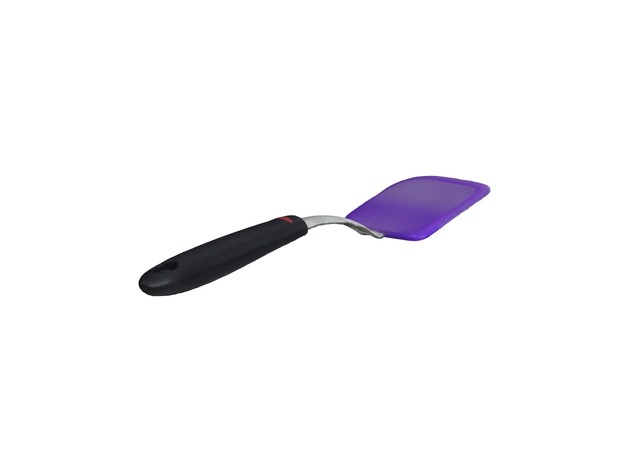} & Spatula with Purple Head: Semantic distractor for purple block. \\
  \hline
\end{longtable}
\twocolumn

\newpage